\documentclass[preprint,12pt]{elsarticle}

\usepackage{amssymb}
\usepackage{amsmath}
\usepackage{amsthm}

\usepackage{pifont}
\usepackage{longtable}
\usepackage{booktabs}
\usepackage{mathrsfs}
\usepackage[ruled]{algorithm2e}
\usepackage{array}
\usepackage{caption}
\usepackage{multirow}
\usepackage{makecell}
\usepackage{threeparttable} 
\usepackage[colorlinks,
            linkcolor=red,
            anchorcolor=blue,
            citecolor=green
            ]{hyperref}
\usepackage{footnote}
\usepackage{threeparttablex}

\journal{Engineering Applications of Artificial Intelligence}

\begin{document}

\begin{frontmatter}
%\texorpdfstring{TEXstring}{PDFstring}



\title{A novel asymmetrical autoencoder with a sparsifying discrete cosine Stockwell transform layer for gearbox sensor data compression}


\author[inst1]{Xin Zhu}
%\ead{xzhu61@uic.edu}

\affiliation[inst1]{organization={Department of Electrical and Computer Engineering},
    addressline={University of Illinois Chicago}, 
    city={Chicago},
    country={USA}}

\author[inst3]{Daoguang Yang}
%\ead{daoguang.yang@polimi.it}
\author[inst1]{Hongyi Pan}
%\ead{hpan21@uic.edu}
\author[inst3]{Hamid Reza Karimi \texorpdfstring{\corref{1}}{textcorref}}
\ead{hamidreza.karimi@polimi.it}
\author[inst2]{Didem Ozevin}
%\ead{dozevin@uic.edu}
\author[inst1]{Ahmet Enis Cetin \texorpdfstring{\corref{1}}{textcorref}}
\ead{aecyy@uic.edu}
\cortext[1]{Corresponding authors}
%\cortext[cor1]{}
%\corref{*}

\affiliation[inst2]{organization={Civil, Materials and Environmental Engineering},
    addressline={University of Illinois Chicago}, 
    city={Chicago},
    country={USA}}

\affiliation[inst3]{organization={Department of Mechnical Engineering},
    addressline={ Politecnico di Milano}, 
    city={Milan},
    country={Italy}}

\begin{abstract}
The lack of an efficient compression model remains a challenge for the wireless transmission of gearbox data in
non-contact gear fault diagnosis problems. In this paper, we present a signal-adaptive asymmetrical autoencoder with a transform domain layer to compress sensor signals. 
First, a new discrete cosine Stockwell transform (DCST) layer is introduced to replace linear layers in a multi-layer autoencoder. A trainable filter is implemented in the DCST domain by utilizing the multiplication property of the convolution. A trainable hard-thresholding layer is applied to reduce redundant data in the DCST layer to make the feature map sparse. In comparison to the linear layer, the DCST layer reduces the number of trainable parameters and improves the accuracy of data reconstruction.
 Second, training the autoencoder with a sparsifying DCST layer only requires a small number of datasets. The proposed method is superior to other autoencoder-based methods on the University of Connecticut (UoC) and Southeast University (SEU) gearbox datasets,  as the average quality score is improved by 2.00\% at the lowest and 32.35\% at the highest with a limited number of training samples.
\end{abstract}


% \begin{highlights}
% \end{highlights}

\begin{keyword}
Gearbox sensor data compression \sep Autoencoder \sep Discrete cosine Stockwell transform \sep Transform domain layer \sep Limited samples
% %% PACS codes here, in the form: \PACS code \sep code
% %% MSC codes here, in the form: \MSC code \sep code
% %% or \MSC[2008] code \sep code (2000 is the default)
\end{keyword}

\end{frontmatter}


\section{Introduction}
\label{sec:Introduction}
The industrial 4.0 paradigm provides an opportunity to use the Internet of Things (IoT) devices to record and transmit condition-monitoring data \cite{compare2022general}. The vibration sensor-based methods are the most popular approaches in the field of rotating machinery fault diagnosis \cite{jiao2022cycle} similar to some traditional fault diagnosis models \cite{gao2015survey, simani2003model} and data-driven fault diagnosis models \cite{zhang2020new, yang2021residual}. It is possible to identify the faults %and Using the vibration signals of the rotating machinery is easy to identify the healthy state of the machine 
\cite{prosvirin2022intelligent} and predict the remaining useful life of a machine from the vibration signals \cite{lv2022vibration}. However, there are still some essential industrial problems that need to be addressed in the field of condition monitoring of rotating machinery. %For example, 
It is not easy to transmit raw condition-monitoring sensor data through wireless networks \cite{huang2015divide} because of high bandwidth requirements. 

To solve the wireless data transition problems, some %traditional 
signal processing techniques are applied to reduce the size of vibration signal data during the transmission process, such as empirical mode decomposition (EMD) \cite{guo2013novel} and singular value decomposition
(SVD) \cite{de2015data}. However, the above-mentioned traditional compression methods can not achieve the balance between compression efficiency and reconstruction accuracy during transmission. Additionally, they are not as effective as the lossy speech, audio, and image compression methods successfully used in wireless communications.
%Moreover, in the field of rotating machinery condition monitoring, there is no effective lossy compression method for gearbox vibration signals based on traditional signal processing approaches. 

With the development of unsupervised learning algorithms, the autoencoder (AE) has been widely used in addressing data compression \cite{sunil2021bio}, dimensional reduction \cite{lu2020multi}, and feature extraction problems \cite{wang2020deep,zemouri2023hydrogenerator}. Hence, it is possible to apply powerful AE to compress gearbox vibration signals, especially its variants: the sparsity AE \cite{ko2022new}. 
However, there are some defects with the existing AE structure in the compression process for gearbox vibration signals. For instance, 
deep convolutional or stacked autoencoders only retain a small proportion of hidden features to represent the original signal by adding many convolution layers in the encoder~\cite{yildirim2018efficient}. But for gearbox data compression tasks, a small number of latent space coefficients limit the accuracy of the reconstruction signal due to the complexity of gearbox data.
Moreover, the utilization of numerous linear or convolutional layers in the sensor introduces significant computational burdens and hardware expenses. Thus, classical deep autoencoders with many encoder layers are unsuitable for gearbox data compression.
% However, there are some defects with the existing AE structure in the compression process for gearbox vibration signals \cite{wang2020novel}, for instance, the model size, compression efficiency, and reconstruction error.

 Neural network-based image compression systems have recently been developed \cite{yue2015beyond}. They produce better results than the most widely used joint photographic experts group (JPEG) standard \cite{wallace1991jpeg} and modern engineered codecs such as the wavelet transform-based JPEG2000 \cite{taubman2002jpeg2000} and the JPEG simulation network (JSNet) \cite{chen2020jsnet}. However, there is no prior work on neural network-based gearbox sensor data compression methods to the best of our knowledge. 

In signal processing and information theory, transform-based methods for data compression have attracted a great deal of attention in the past few decades~\cite{taubman2002jpeg2000,aydin1991ecg}. Among these, discrete cosine Stockwell transform (DCST) has excellent compression performance. The authors in \cite{jha2018electrocardiogram} have taken advantage of the orthogonality of transforms including discrete cosine transform (DCT), and DCST to reduce redundant data for electrocardiogram (ECG) data compression ~\cite{243411,aydin1991ecg,cetin2006compression,cetin1994coding}. It has also been shown that DCST has better energy concentration properties than DCT in some image compression experiments \cite{978}. DCST uses DCT to divide the data into subbands similar to the audio compression methods therefore it is more suitable for streaming gearbox sensor data compression than a straightforward application of the DCT onto the gearbox data.

In this paper,
a novel asymmetrical autoencoder with a transform domain layer is proposed for the lossy compression of gearbox sensor signals. 

The proposed autoencoder performs data compression similar to the transform domain coding algorithms which are successfully used in audio, image, and video coding.
It combines fully connected layers with a DCST-based layer to obtain an architecture whose latent-space thresholding and frequency domain scaling parameters can be optimized using a gradient-based algorithm.
The DCST imposes a fixed structure in the encoder to cover all the frequency components of the gearbox sensor signal and it prevents over-smoothing. 
%is implemented in the encoder to compress the gearbox sensor signal. 
Additionally, the sparsity penalty-based cost function is used to improve the performance of the hard-thresholding module in the DCST layer to obtain a sparse latent space vector.
Then the DCST coefficients are transmitted or stored using entropy coding. Finally, the received signal is recovered using inverse DCST and a fully connected linear layer. The main contributions of this paper are concluded as follows:
\begin{itemize}

\item A new DCST layer is proposed to replace linear layers. Trainable quantization parameters and hard thresholds are introduced into the DCST layer to decrease the trainable parameters and improve the compression ratio by zeroing out small valued DCST coefficients. 

\item A computationally efficient asymmetrical autoencoder is developed for gearbox sensor data compression. The encoder module has low complexity, making it suitable for implementation in a low-cost sensor. The network is trained offline using the backpropagation algorithm.

\item The online data transmission module can achieve the balance between compression efficiency and reconstruction accuracy. Experimental results demonstrate a good compression performance on two gearbox datasets: the SEU~\cite{shao2018highly} and the UoC gearbox datasets \cite{cao2018preprocessing}. 
\item The goal of this paper is not to develop a new gear fault identification method. 
However, it is experimentally shown that the proposed model performs efficient gearbox data compression without sacrificing any useful information for fault identification.

\end{itemize}

\section{Preliminaries}
\label{sec:Background}
This section describes related autoencoder methods, discrete cosine transform (DCT), and transform domain data compression, whose one of the earliest versions is the well-known JPEG standard, which uses the DCT as the main building block. \cite{wallace1991jpeg}. 
 Our neural network-based data compression method is not a standard autoencoder, however, it embeds DCT into the well-known autoencoder structure to take advantage of both deep neural networks and classical data compression methods.
We first review the autoencoder structure.
\begin{figure}[htbp]
	\centering
		\includegraphics[scale=.13]{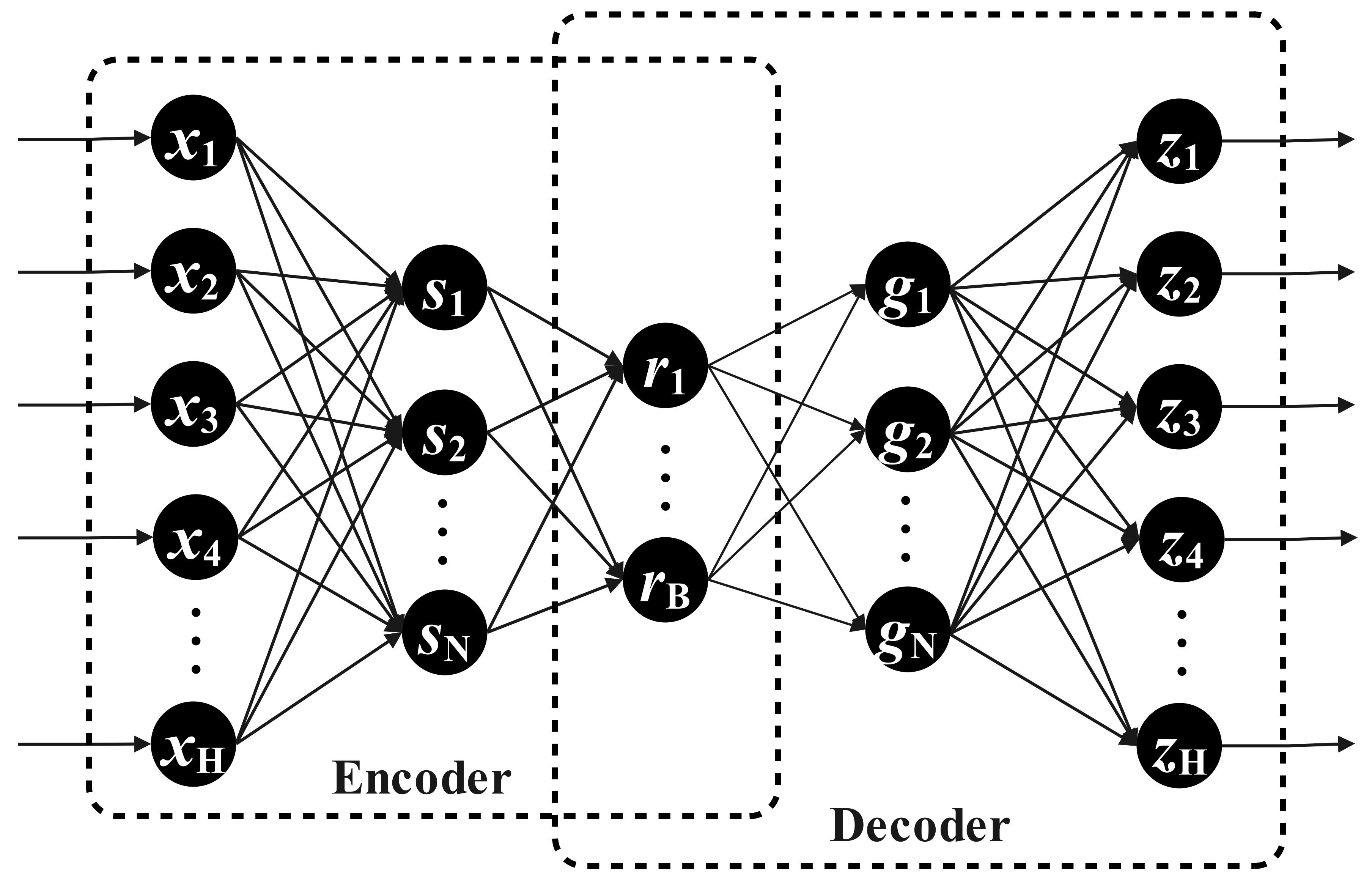}
	\caption{Multi-layer autoencoder structure}
	\label{Fig:AE}
\end{figure}
\subsection{ Multi-layer autoencoder}
The multi-layer autoencoder is a type of neural network that extracts the hidden features from the input data~\cite{tan2008performance}. As shown in Fig.\ref{Fig:AE}, the multi-layer autoencoder structure contains two parts: an encoder and a decoder. The encoder includes an input layer and several fully connected layers. The decoder is composed of an output layer and several fully connected layers. The activation function such as the sigmoid (or rectified linear unit (ReLU)) is followed by a fully connected layer. 
The fully connected layer processing from $\mathbf{x}\in\mathbb{R}^{\Omega}$ to $\mathbf{y}\in\mathbb{R}^{\Psi}$ is computed as:
\begin{equation}\label{Eq:linear}
    \mathbf{y} = \delta(\mathbf{Wx}+\mathbf{b}),
\end{equation}
where, $\delta(\cdot)$ stands for the activation function, $\mathbf{W}\in\mathbb{R}^{\Psi\times \Omega}$ is the weight coefficient matrix and $\mathbf{b}\in \mathbb{R}^{\Psi}$ is the bias term.

During the training process, the autoencoder minimizes the loss function to find the optimal weight coefficients and biases. The loss function is calculated as:
\begin{eqnarray}\label{Eq:Loss_linear}
Loss=\frac{1}{H}\sum_{i=0}^{H-1} ({x}_{i}-{z}_{i})^2,
\end{eqnarray}
where ${x}_{i}$ is the $i$-th input neuron and ${z}_{i}$ is the corresponding output neuron of the autoencoder, respectively; $H$ is the number of neurons in the input layer.

The encoder part of the network usually transforms the input data into a lower dimensional space and this lower dimensional representation can be used to compress the input data. Therefore, the autoencoder is a nonlinear version of the Karhunen-Loeve transform (KLT). KLT-based methods \cite{243411,cetin2006compression} compute the eigenvectors of the covariance matrix of the input similar to the principal component analysis (PCA) and use the significant eigenvectors to represent the data in a lower dimensional subspace \cite{ahmed1974discrete}.

\subsection{Discrete Cosine Transform}
The discrete cosine transform (DCT) approximates the KLT when there is a high correlation among the input samples~\cite{ahmed1974discrete}. The DCT-based methods represent a given signal in a lower dimensional latent space, which makes it better suited to compression similar to the KLT without requiring the covariance matrix estimation and eigen-analysis.
For a vector of $N$ real numbers $\{c_0,c_1,\dots,c_{N-1}\}$, an orthogonal type-II DCT coefficients $C_k$ is defined as:
% \begin{equation}
%     C_k = \sum_{n=0}^{N-1}c_n\text{cos}\left[\frac{\pi}{N}\left(n+\frac{1}{2}\right)k\right],\label{eq: DCT}
% \end{equation}
% for $0\leq k \leq N-1$. 
% If it is an orthogonal DCT, $ c_k$ should be multiplied by a scaling factor $f$:
\begin{equation}\label{Eq:dct}
C_k=\left\{
\begin{aligned}
&\sqrt{\frac{1}{N}} \sum_{n=0}^{N-1}c_n,  &k=0,\\
&\sqrt{\frac{2}{N}} \sum_{n=0}^{N-1}c_n\text{cos}\left[\frac{\pi}{N}\left(n+\frac{1}{2}\right)k\right], &k > 0,\\
\end{aligned}
\right.
\end{equation}
for $0\leq k \leq N-1$.
The orthogonal IDCT is given by :
\begin{equation}
    \hat{c}_k=\sqrt{\frac{1}{N}} C_0 +\sqrt{\frac{2}{N}}\sum_{n=1}^{N-1}C_n\text{cos}\left[\frac{\pi}{N}\left(k+\frac{1}{2}\right)n\right].\label{eq: idct}
\end{equation}

The DCT is used in audio, video, and image compression standards such as the JPEG compression standard which is a classic image compression method \cite{wallace1991jpeg}. 
Similar to the transform domain data compression methods, we divide the input 
into small blocks of data and compute the discrete cosine Stockwell transform (DCST) of each block after a fully connected neural network layer. 
Specifically, the input data is represented by a smaller set of DCST coefficients. The details of DCST will be described in subsection~\ref{sec:Discrete}. 
%Then, it is experimentally observed that DCST produces better results than DCT.

\subsubsection{Quantization and dequantization}
\begin{figure}[htbp]
	\centering
		\includegraphics[scale=.37]{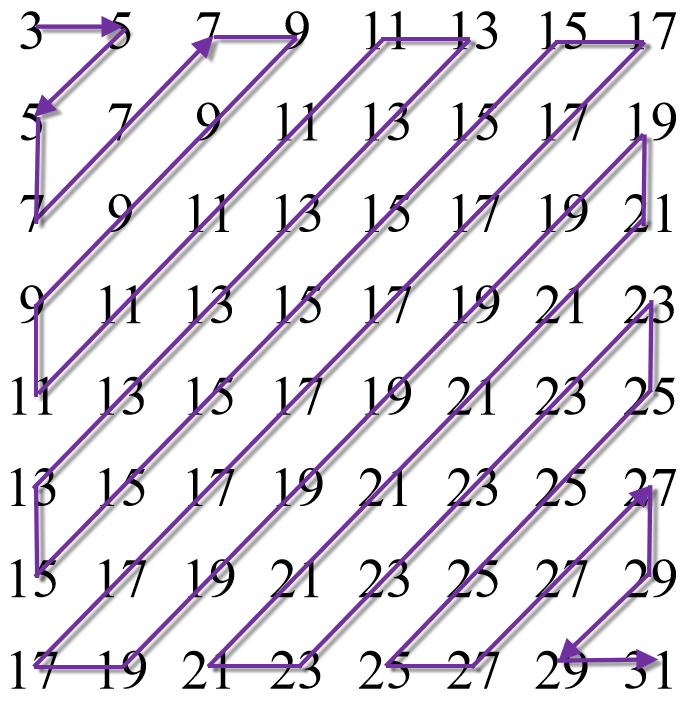}
	\caption{A Sample Quantization Matrix}
	\label{Fig:zigzag}
\end{figure}

%One of the important features of the JPEG algorithm is that it uses a quantization matrix.
After the block DCST, we quantize the data 
using a quantization matrix similar to the JPEG algorithm. In the JPEG algorithm, the two-dimensional (2D) DCT coefficient matrix $\mathbf{F}$ is element-wise divided by a quantization matrix $\mathbf{G}$ and rounded as follows:
\begin{eqnarray}\label{Eq:Round}
{F}^{\mathbf{G}}(u,v)=\text{Round}\left(\frac{F(u,v)}{G(u,v)}\right),
\end{eqnarray}
where $F(u,v)$ represents the $(u,v)-th$ 2D DCT coefficient of the DCT matrix $\mathbf{F}$ , and $G(u,v)$ is the $(u,v)-th$ element of the quantization matrix $\mathbf{G}$, respectively, and
$\text{Round}(\cdot)$ is the integer rounding operation. Transform domain parameters are scaled using Eq.~(\ref{Eq:Round}).
Rounding operation is a lossy operation and it discretizes (quantizes) the transform domain coefficients.
Elements $G(u,v)$  of the quantization matrix $\mathbf{G}$ are hand-crafted by experts. A sample quantization matrix~\cite{qtable} is shown in Figure \ref{Fig:zigzag}. 
Usually, more emphasis is given to "low-frequency" coefficients compared to high-frequency coefficients. 

 The dequantization is performed as the element-wise product $\circ$ between $\mathbf{F}^{\mathbf{G}}$ and $\mathbf{G}$:
\begin{eqnarray}\label{Eq:Deq}
\widetilde{\mathbf{F}}= \mathbf{F}^{\mathbf{G}} \circ \mathbf{G}.
\end{eqnarray}

% In our method, we will also use a quantization matrix (or scaling matrix) in the transform domain similar to the JPEG algorithm. We learn the parameters of the quantization matrix using the backpropagation algorithm during training but we initialize our parameters similar to the quantization matrix of the JPEG standard, i.e., we give less emphasis to high-frequency coefficients.
%\subsubsection{Entropy coding and entropy decoding}
After quantization, entropy coding is applied as the final step in many lossy data compression schemes including the JPEG algorithm to convert the data to a bitstream. This step is lossless and converts the quantized transform domain coefficients to a data stream. Before the entropy coding, the elements in the coefficient matrix need to be rearranged into a one-dimensional sequence in the zigzag order as shown in Fig.~\ref{Fig:zigzag}.

\section{Framework of the proposed method}
\label{sec:Methods}

\subsection{Motivation}
Gear vibration signals can be divided into two categories: normal and faulty gear vibration signals. The gear in normal operation also generates vibration during transmission due to its own stiffness. Its waveform is a periodic attenuation waveform, and its low-frequency signal has a meshing waveform similar to a sine wave. However, in faulty cases, the sine wave meshing waveform will be corrupted by missing teeth, tooth root fault, surface fault, etc. The corruption makes it difficult to compress the faulty gear data directly. 

At present, autoencoders have been widely applied to many other compression tasks~\cite{wang2019novel}. Classical autoencoders have a symmetric U-net type structure in which both the encoder and decoder sides have an equal number of linear layers. However, the encoder side must contain a large number of trainable parameters to obtain the latent representation of a given signal. Too many fully connected layers in the encoder or decoder will also make the model easily overfit the training data. In addition, a large number of linear or convolution layers will bring significant computational costs to the sensor, along with increased transmission delay and hardware costs.

\subsection{Asymmetrical autoencoder with a  sparsifying DCST layer}
\begin{figure*}[htbp]
	\centering
		\includegraphics[scale=.14]{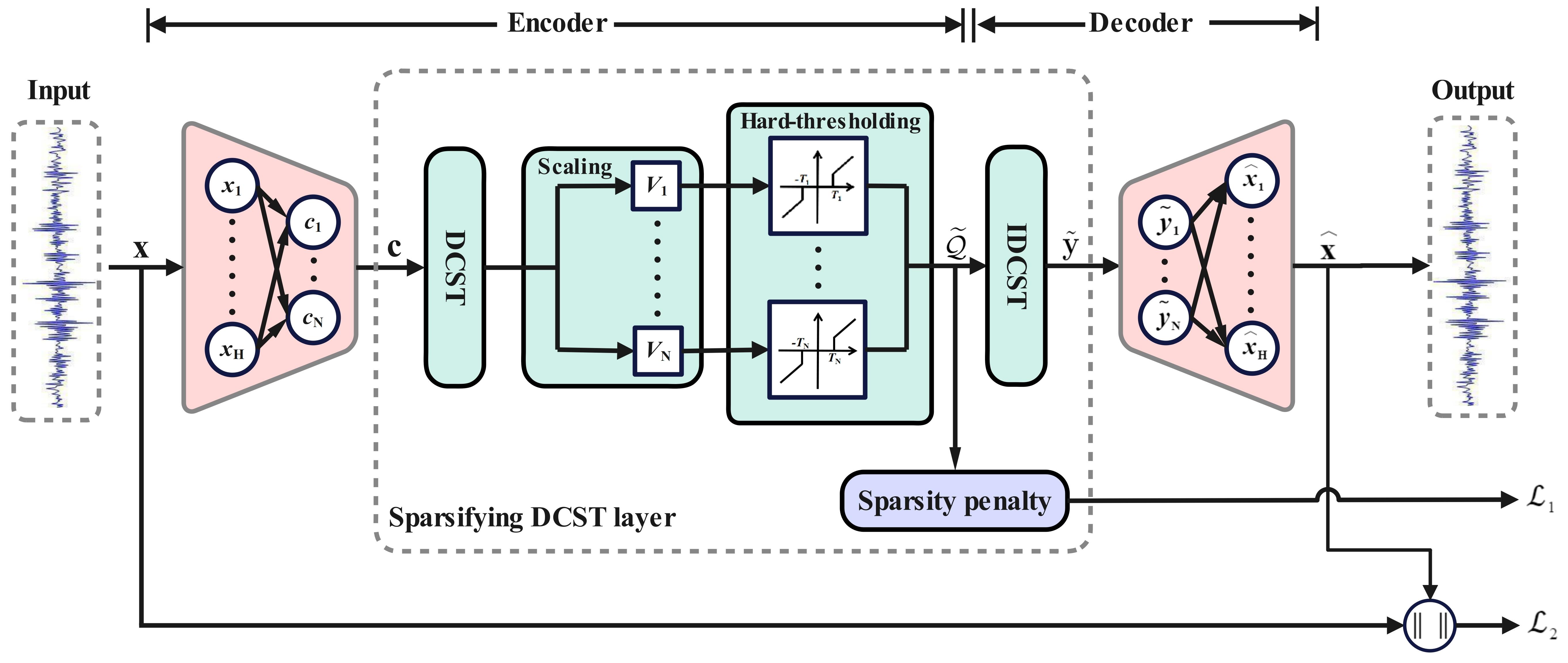}
	\caption{Asymmetrical autoencoder with a  sparsifying DCST layer.}
	\label{Fig:AE_DCST}
\end{figure*}

Transform domain methods including the image compression method JPEG and the video compression standards MPEG-II and MPEG-IV etc. are the most prevalent image and video data compression methods in the last three decades~\cite{chen2021feature,618009}. Inspired by the above-mentioned transform domain methods, the  DCST layer for gearbox data compression is incorporated into the autoencoder network structure in this work. The framework of the proposed sparsifying DCST layer is illustrated in Fig.~\ref{Fig:AE_DCST}. 
We will use the DCST method to transform the autoencoder feature map into the transform domain. 
Our proposed autoencoder system learns the scaling (quantization matrix) parameters and thresholds from the gear data in an automatic manner to achieve a sparse and lower dimensional representation of the original signal in the latent space. 
After that, the system calculates the sparsity penalty during the training process. Then, the inverse DCST (IDCST) is applied to reconstruct the data.  

The proposed asymmetrical autoencoder structure with a sparsifying DCST layer has fewer trainable parameters than a regular symmetrical autoencoder with four linear layers shown in Fig.~\ref{Fig:AE}. In our network, the DCST layer replaces the middle two linear layers of the autoencoder, i.e., the encoder part of our structure has one linear layer, DCST, scaling layer, and hard-thresholding layer as shown in Fig.~\ref{Fig:AE_DCST}. The decoder part reconstructs the transform domain data using IDCST and linear layer.

The input sensor signal is divided into $H$-length short-time windows (blocks). Then, the autoencoder processes the data block by block. 
As shown in Fig.~\ref{Fig:AE_DCST},
the input vector $\mathbf{x}\in\mathbb{R}^{H}$ is first processed by a linear layer. 
%From 
We considered two cases for the activation function $\delta(\cdot)$ of the linear layer (Eq.~(\ref{Eq:linear})) in our experimental studies.
The output after the first linear layer is
\begin{eqnarray}\label{Eq:first_out}
\mathbf{c}=\tanh{\left(\mathbf{W}^{(1)}\mathbf{x}+\mathbf{b}^{(1)}\right)},
\end{eqnarray}
where $\mathbf{W}^{(1)}$ and $\mathbf{b}^{(1)}$, represent the weights matrix and biases vector of the first layer.
In a completely linear layer where $\delta$ is the identity function, we have, 
\begin{eqnarray}\label{Eq:first_out2}
\mathbf{c}=\left(\mathbf{W}^{(1)}\mathbf{x}+\mathbf{b}^{(1)}\right).
\end{eqnarray}

%$\mathbf{c}=\left(\mathbf{W}^{(1)}\mathbf{x}+\mathbf{b}^{(1)}\right)$.
% Then, the DCST layer calculates the DCST of the vector $\mathbf{c}$ and obtains the vector $\tilde{\mathbf{y}}$. If $N\leq 64$, zeros are padded after the input tensor.

\subsubsection{Discrete Cosine Stockwell Transform}
\label{sec:Discrete}
\begin{figure}[htbp]
	\centering
		\includegraphics[scale=.4]{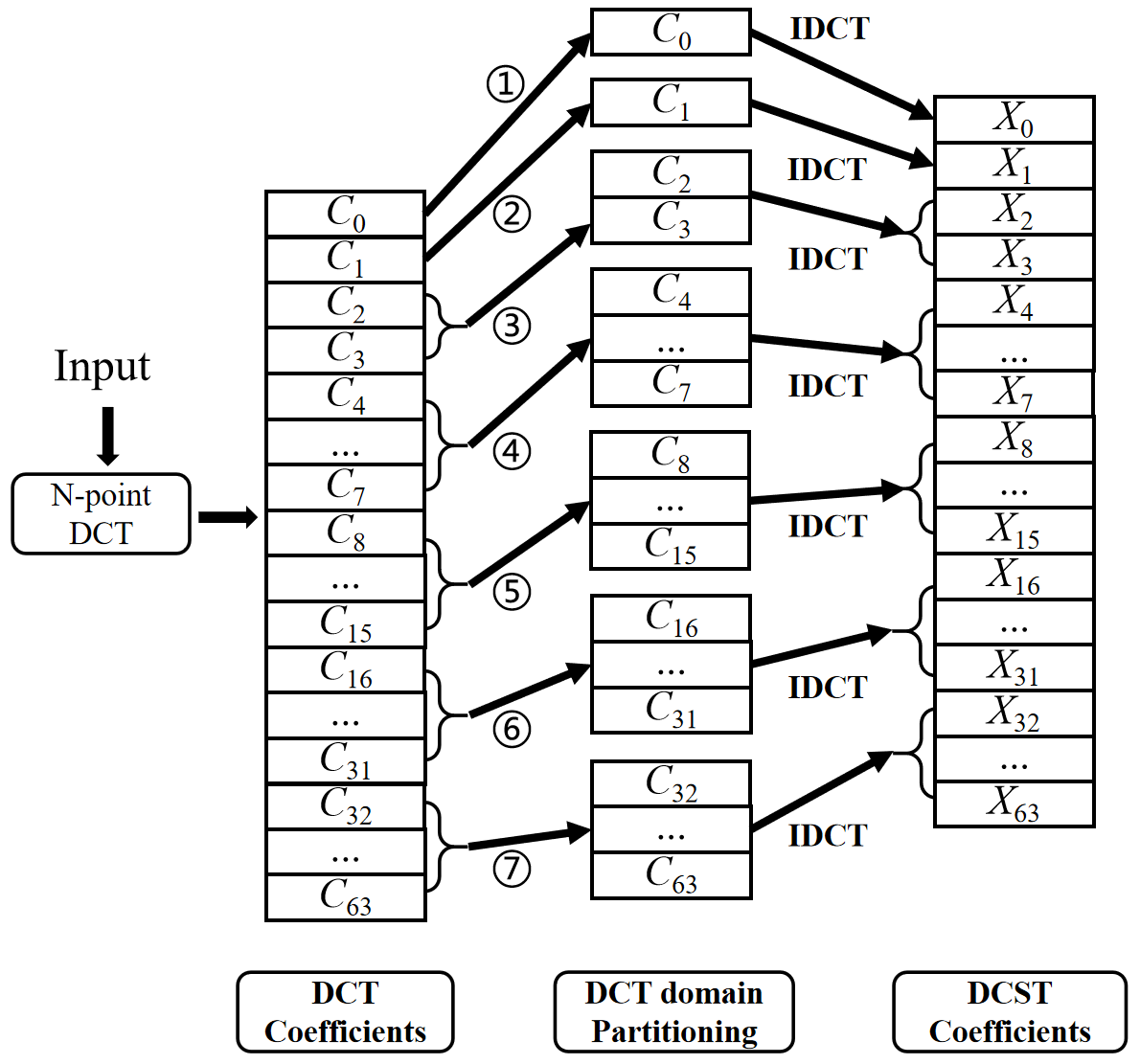}
	\caption{Block diagram of 64 point-DCST implementation.}
	\label{Fig:DCST}
\end{figure}

In this work, the orthogonal type-II DCT (Eq.~(\ref{Eq:dct})) is utilized in the discrete cosine Stockwell transform (DCST)~\cite{jha2018electrocardiogram}. DCST acts like a wavelet filter bank and divides the DCT coefficients into frequency bands. DCT coefficients do not perfectly represent the frequency content of the input but they have frequency content \cite{ahmed1974discrete}. For a given signal $\mathbf{c}$ with the length of $N = 2^M, M\in\mathbb{N}$, the DCST is computed as shown in Fig.~\ref{Fig:DCST}. 
First, the DCT of the current input block (or the block of data from the previous layer) is computed. 
In this paper, the DCT size is experimentally selected as $N=64$. The resulting coefficients  $\mathbf{C}=[C_0,C_1,\dots, C_{63} ]^{\mathrm{T}}$ are divided into seven frequency sub-bands. The length of each frequency sub-band is given by $\{n_{i}|1\leq i\leq 7\}$; $n_{i}$ represent the width of the frequency partitions. More specifically,
\begin{equation}
\begin{aligned}
&n_{1}=1,\\
&n_{i}=2^{i-2},2\leq i\leq 7.\\
\end{aligned}
\end{equation}
%where $M=log_2 (N)$.
Then $n_{i}$ point inverse DCT is applied to the corresponding frequency sub-band. 
After that, $X_{0},X_{1},\dots,X_{N-1}$ are treated as the DCST coefficients. Furthermore, feature coefficients are formed by concatenating and rearranging the DCST coefficients.
The DCST is summarized in Algorithm \ref{algorithm:DCST}.
% \begin{equation}
%     \mathbf{DCST}=\left(\overset{K-1}{\underset{i=0}{\biguplus}}\mathbf{DCT}_{n_i}^{-1}\right)\mathbf{DCT}
%     ,\label{Eq:DCST}
% \end{equation}
% where $\mathbf{DCT}^{-1}$ is inverse DCT. $\biguplus$ represents the concatenation of the frequency partitions. 
The input data can be perfectly reconstructed from DCST coefficients in the absence of any quantization because the DCT is an orthogonal and invertible transform. 
\begin{algorithm}
 %\SetAlgoNoLine  %
 \caption{DCST}
 \label{algorithm:DCST}
  \KwIn{Input tensor ${\mathbf{c}} \in\mathbb{R}^{N}$}
  \KwOut{Output tensor ${\mathbf{X}} \in\mathbb{R}^{N}$}
        ${\mathbf{C}}={\mathbf{DCT}}(\mathbf{c})\in\mathbb{R}^{N}$; \\
        \For{$i=2;i \le \log_{2}N+1;i=i+1$}
        {
            $\beta=2^{i-2}$; \\
            $\lambda=\beta-1$;\\
            $\mathbf{X}[\beta:\lambda+\beta]={\mathbf{IDCT}}(\mathbf{C}[\beta:\lambda+\beta])\in\mathbb{R}^{N}$; \\
        }
        return ${\mathbf{X}}\in\mathbb{R}^{N}$; \\
\end{algorithm}
\subsubsection{Scaling and hard-thresholding}
Unlike the hand-crafted quantization schemes used in JPEG and MPEG, a trainable quantization vector $\mathbf{V}$ is used during the quantization process in the transform domain. Each DCST coefficient $X_{i}$ is element-wise divided by the trainable weight parameter $V_{i}$:
\begin{eqnarray}\label{Eq:scaling}
\mathbf{\mathcal{Q}}_{i}=\frac{X_{i}}{V_{i}},
\end{eqnarray}
for $0 \leqslant i \leqslant N-1$. Each weight parameter $V_{i}$ is learned using the backpropagation algorithm. We call this step the scaling (filtering) operation \cite{pan2022dct}. 
We initialized the scaling parameters using the weights 
shown in Fig. \ref{Fig:zigzag}. We zig-zag scanned the quantization elements of the JPEG quantization matrix and rearranged them into a vector $\mathbf{q}=[q_{0}\ q_{1}\ \ldots\ q_{63}]^{\mathrm{T}}$. Then $\mathbf{q}$ is applied to initialize the  weight vector $\mathbf{V}= [V_{0}\ V_{1}\ \ldots\ V_{63}]^{\mathrm{T}}$ which is trained using the backpropagation algorithm.

Another interpretation of the scaling layer is related to the multiplication property of the convolution. In type-II DCT\cite{park2003m}, element-wise multiplication in the time domain corresponds to symmetric convolution in the DCT domain.  It can be written as:
\begin{equation}
    \mathbf{f}\ \circ\ \mathbf{g} = \mathcal{D}^{-1}(\mathbf{\Gamma}\circ((\mathbf{\Phi}\circ\mathcal{D}(\mathbf{f}))\otimes_s(\mathbf{\Phi}\circ\mathcal{D}(\mathbf{g}))))
\end{equation}
where $\circ$ represents the elementwise multiplication, $\mathbf{f}, \mathbf{g}\in\mathbb{R}^N$; $\mathcal{D}(\cdot)$ stands for an orthogonal type-II DCT; $\mathcal{D}^{-1}(\cdot)$ stands for an orthogonal type-II IDCT; $\otimes_s$ stands for the symmetric convolution;
%and $\circ$ stands for the element-wise product.
$\mathbf{\Gamma}$ is a constant vector and $\mathbf{\Phi}$ is given as
%$\mathbf{\Gamma}[i]=\frac{1}{\mathbf{\Phi}[i]}, 0<i<m$. 
\begin{equation}\label{Eq:coe}
\mathbf{\Phi}_i=\left\{
\begin{aligned}
&2\sqrt{{N}},  &i=0,\\
&\sqrt{2N} , &i > 0,\\
\end{aligned}
\right.
\end{equation}
where $0\leq i\leq N-1$. 
Therefore, for the $t$-th frequency sub-band of DCST, we have
\begin{equation}
    \mathbf{X}_t\circ {\mathbf{V}_t^{-1}}=\mathcal{D}^{-1}(\mathbf{\Gamma}\circ((\mathbf{\Phi}\circ\mathcal{D}(\mathbf{X}_t))\otimes_s(\mathbf{\Phi}\circ\mathcal{D}({\mathbf{V}_t^{-1}})))),             
\end{equation}
where $1\leq t\leq 7$ and
\begin{equation}\label{Eq:X}
\mathbf{X}_{t}=\left\{
\begin{aligned}
&[X_{0}],  &t=0,\\
&[X_{2^{t-1}} \ldots\ X_{2^{t}-1}]^{\mathrm{T}} , &t > 0,\\
\end{aligned}
\right.
\end{equation}
\begin{equation}\label{Eq:V}
\mathbf{V}_{t}^{-1}=\left\{
\begin{aligned}
&[1/ V_{0}],  &t=0,\\
&[1/ V_{2^{t-1}} \ldots\ 1/V_{2^{t}-1}]^{\mathrm{T}} , &t > 0.\\
\end{aligned}
\right.
\end{equation}
% where $\mathbf{V}_{t}^{-1}=[\frac{1}{V_{0}}\ \frac{1}{V_{1}}\ \ldots\ \frac{1}{V_{63}}]^{\mathrm{T}}$.

Next, the DCT of the $({\mathbf{X}_t}\circ{\mathbf{V}_t^{-1}})$ is computed as follows:
\begin{equation}
    \mathcal{D}(\mathbf{X}_t\circ {\mathbf{V}_t^{-1}})=\mathbf{\Gamma}\circ((\mathbf{\Phi}\circ\mathcal{D}(\mathbf{X}_t))\otimes_s(\mathbf{\Phi}\circ\mathcal{D}({\mathbf{V}_t^{-1}}))),             \label{formula}
\end{equation}

As shown in Fig.~(\ref{Fig:DCST}), $\mathbf{X}_t=\mathcal{D}^{-1}(\mathbf{C}_t)$. Then, Eq.~(\ref{formula}) can be converted to
\begin{equation}
    \mathcal{D}(\mathbf{X}_t\circ {\mathbf{V}_t^{-1}})=\mathbf{\Gamma}\circ((\mathbf{\Phi}\circ\mathbf{C}_t)\otimes_s(\mathbf{\Phi}\circ\mathcal{D}({\mathbf{V}_t^{-1}}))),       
\end{equation}
where
\begin{equation}\label{Eq:C}
\mathbf{C}_{t}=\left\{
\begin{aligned}
&[C_{0}],  &t=0,\\
&[C_{2^{t-1}} \ldots\ C_{2^{t}-1}]^{\mathrm{T}} , &t > 0.\\
\end{aligned}
\right.
\end{equation}

\begin{figure}[htbp]
	\centering
		\includegraphics[scale=.15]{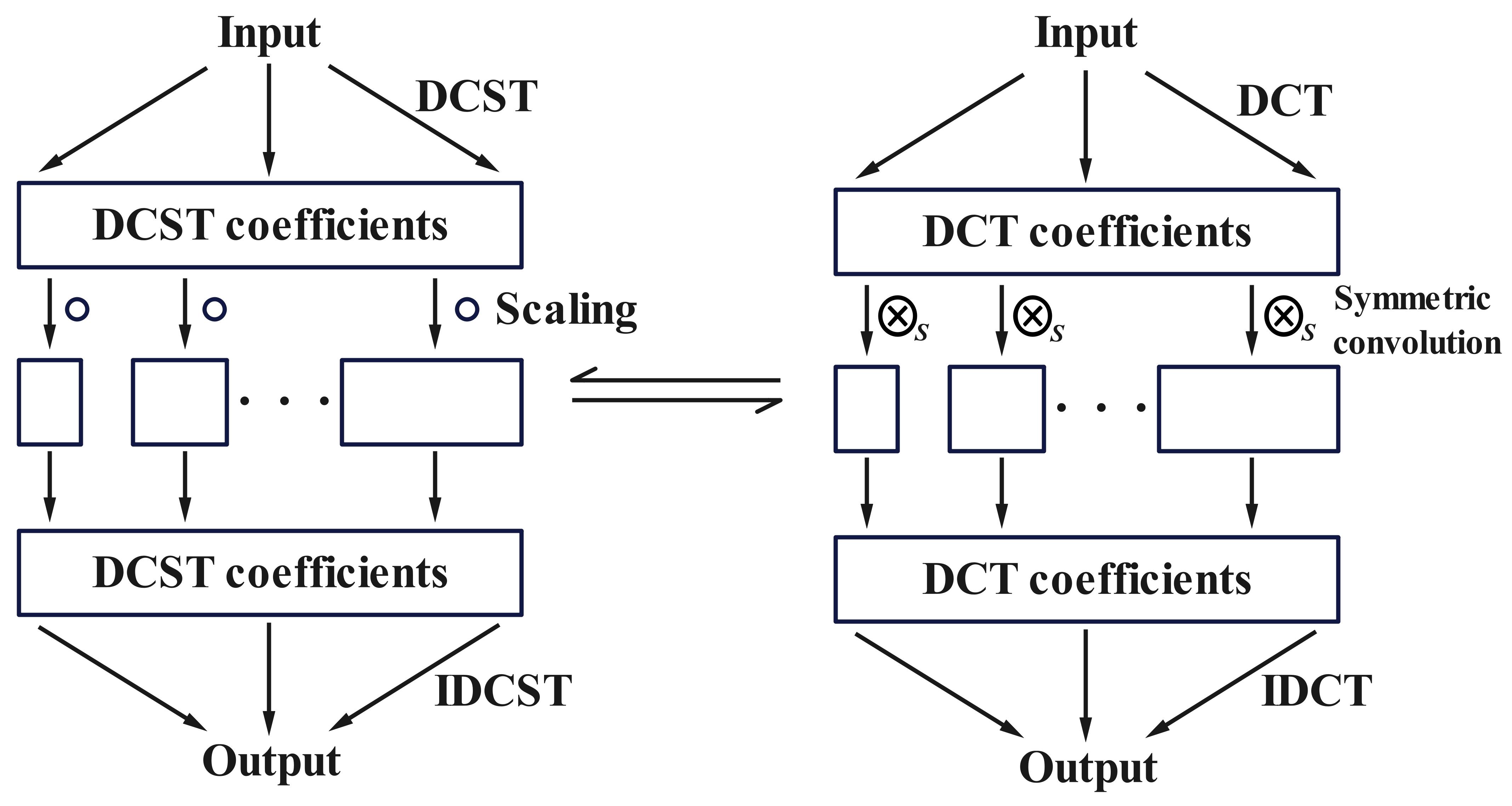}
	\caption{Relationship between scaling operation in the DCST domain and convolution in the DCT domain.}
	\label{Fig:compare}
\end{figure}

Therefore, the scaling operation in the DCST domain is similar to band-pass filtering because it can be converted into symmetric convolution in the DCT domain as shown in Fig.~\ref{Fig:compare}. The authors in~\cite{jiang2019dct} show that convolution in the DCT domain can effectively extract important features. 

 There are seven frequency sub-bands and scaling sequences in the DCST domain as shown in Fig.~\ref{Fig:DCST}. Each scaling sequence can be treated as a special convolution kernel, and the combination of seven convolution kernels can be regarded as a filter bank. Compared with direct utilization of the scaling layer in the DCT domain, a filter bank can extract temporal dependency features and improve the reconstruction accuracy compared to regular DCT-based data compression. 
Furthermore, it is experimentally observed that the scaling operation produces better results in the DCST domain than in the DCT domain in Subsection~\ref{sec:Comparison}. 
Hence, it can be concluded that the trainable scaling layer in the DCST domain can scale the DCST coefficients to improve the compression ratio. Additionally, it works as a filter bank in the DCT domain to take advantage of temporal dependencies. %features.
% there are seven frequency sub-bands and scaling sequences in the DCST domain as shown in Fig.~\ref{Fig:DCST}. Each scaling sequence can be treated as a special convolution kernel in the DCT domain. A combination of seven convolution kernels can be regarded as a filter bank in the DCT domain. 
% Compared with directly utilizing the scaling layer in the DCT domain, a wavelet
% filter bank can extract more important features and improve the reconstruction accuracy using the convolution property. 

Next, instead of using the integer rounding operation, a trainable hard-thresholding layer is applied to remove the small entries in the DCST domain similar to image coding and transform domain denoising \cite{mallat1999wavelet}. Hard-thresholding has the advantage of not changing the value of large entries and retains the energy of the input more compared with soft-thresholding~\cite{donoho1995noising}.
We determine the thresholds using the backpropagation algorithm. The hard-thresholding is defined as:
\begin{equation}
    \begin{split}
        \widetilde{{\mathcal{Q}}}_i= \mathcal{S_T}\left({\mathcal{Q}}_{i}\right)+T_i\cdot\text{sign}\left(\mathcal{S_T}\left({\mathcal{Q}}_{i}\right)\right),
    \end{split}
    \label{Eq:HT}
\end{equation}
where, %$\mathcal{S_T}(\tilde{X}_i)= \text{sign}(\tilde{X}_i)(|\tilde{X}_i|-T_i)_{+}$. 
\begin{equation}
    \mathcal{S_T}\left({\mathcal{Q}}_{i}\right)= \text{sign}\left({\mathcal{Q}}_{i}\right)\left(|{\mathcal{Q}}_{i}|-T_i\right)_{+}
\end{equation}
is the soft-thresholding function;
$T_i$ is a trainable threshold parameter for $i=0, 1,...,63$, and $(\cdot)_+$ is the ReLU function. Hard-thresholding functions are depicted in Fig. \ref{Fig:AE_DCST}.
The hard-thresholded DCST coefficients are referred to as the latent space representation of the original sensor data. 

% \textcolor{blue}{
% The computational cost of the fast DCT algorithm is Order($N\log_2 N$). The DCST can be also implemented using the fast DCT and its computational cost is 
% %Cost:
% %Normal DCST:  $4/3+(8/3)N^2-2N$
% $\frac{5}{2}N\log_2 N-3N+6$.
% }

In contrast to the JPEG algorithm, inverse scaling is not required in the decoder side of the DCST layer because the linear layer after the inverse DCST learns the weights during training and it automatically compensates for the effect of inverse scaling.
%the scaling vector $\mathbf{V}$ is trainable. 
Additionally, with the sparsity penalty, the scaling and hard-thresholding layer can eliminate redundant data and retain important components. The overall operation can be summarized as follows:
\begin{equation}
   \tilde{\mathbf{y}}= \mathscr{D}^{-1}(\mathcal{H}_{T}(\mathscr{D}(\mathbf{c})\circ \mathbf{V}))
\end{equation}
where $\mathcal{H}_{T}(\cdot)$ stands for the hard-thresholding operation; $\mathscr{D}(\cdot)$ stands for DCST and $\mathscr{D}^{-1}(\cdot)$ stands for the inverse DCST.  

%RED alarm or red alert.

After this stage,  the output of the network is obtained
using a linear layer as follows:
\begin{eqnarray}\label{Eq:out}
\mathbf{\hat{x}}=\mathbf{W}^{(2)}\tilde{\mathbf{y}}+\mathbf{b}^{(2)}
\label{wholeoutput}
\end{eqnarray}
where $\mathbf{W}^{(2)}$ and $\mathbf{b}^{(2)}$, represent the weights matrix and biases vector of the last layer at the decoder side as shown in Fig.~\ref{Fig:AE_DCST}.

\subsubsection{Sparsity Penalty Based Cost Function}

To improve the performance of data compression after hard-thresholding, sparsity restrictions~\cite{ng2011sparse} are imposed during training as shown in Fig.~\ref{Fig:AE_DCST}. A space vector in the latent domain with many zeros is more efficient to compress than a vector containing many small values. Let $\widetilde{\mathbf{\mathcal{Q}}}\in\mathbb{R}^N$ represent the output of the hard-thresholding operation. The vector $\widetilde{\mathbf{\mathcal{Q}}}$ is used as a part of the overall loss function of the network after going through a softmax layer.  The activity of the $j$-th component is defined as:
\begin{equation}
    \hat{\gamma}_{j}=\frac{e^{\lvert \widetilde{{\mathcal{Q}}}_j\rvert}}{\sum_{k=0}^{N-1}e^{\lvert \widetilde{{\mathcal{Q}}}_k\rvert}}, j=0,1,\cdots,N-1
\end{equation}
% where $h_{i,j}$ is the $i$-th output of the $j$-th neuron. $L$ is the batch size. $\mathbf{h}_i=\left[\lvert h_{i,0}\rvert\ \lvert h_{i,1}\rvert\ \cdots\ \lvert h_{i,N-1}\rvert\right]^\mathrm{T}$. $N$ is the number of neurons in layer $n$. 
% Further, the average activity of $j$-th neuron in layer $n$ can be defined as:
% \begin{equation}
%     \hat{\gamma}_{j}^{(n)}=\frac{1}{L}\sum_{i=0}^{L-1}a\left(h_{i,j}\right)
% \end{equation}

%The limitation making most of the neurons inhibited is called the sparsity restriction.
%Sparsity is explained as the neuron being considered to be inhibited when $\hat{\mathbf{\gamma}}_{j}$ is close to 0. 
To obtain a sparse $\widetilde{\mathbf{\mathcal{Q}}}$ vector, an additional sparsity penalty factor is added to the loss function. This penalty factor keeps the average activity of neurons in a small range when $\hat{\mathbf{\gamma}}_{j}$ is far away from 0. %There are many reasonable choices for the specific form of penalty factor, and 
In this work, we use the Kullback–Leibler divergence~\cite{ng2011sparse} as the sparsity penalty term, which is defined as
\begin{equation}
    \sum_{j=0}^{N-1} {\rm{KL}}(\gamma\vert\vert\hat{\mathbf{\gamma}}_{j})
    =\sum_{j=0}^{N-1}\gamma\log\frac{\gamma}{\hat{\mathbf{\gamma}}_{j}}+
    (1-\gamma)\log\frac{1-\gamma}{1-\hat{\mathbf{\gamma}}_{j}},
    \label{Eq:KL}
\end{equation}
where ${\gamma}$ is a sparsity parameter, which is a small value close to 0. When $\hat{\mathbf{\gamma}}_{j}={\gamma}$, ${\rm{KL}}\left({\gamma}\vert\vert\hat{\mathbf{\gamma}}_{j}\right)=0$. 
With the increase of $\vert\hat{\mathbf{\gamma}}_{j}-{\gamma}\vert$, $\sum_{j=1}^{N} {\rm{KL}}\left({\gamma}\vert\vert\hat{\mathbf{\gamma}}_{j}\right)$ also increases.

The overall loss function $\mathcal{L}$ of the network is a linear combination of mean square error and the KL divergence.
%defined in Eq. \ref{Eq:KL}.
%as discussed in the next subsection.
It is defined as follows:
\begin{eqnarray}\label{Eq:loss_p}
\mathcal{L}=\mathcal{L}_1+\mathcal{L}_2=\eta\sum_{j=0}^{N-1} {\rm{KL}}\left(\gamma\vert\vert\hat{\mathbf{\gamma}}_{j}\right)+\frac{1}{H}\sum_{i=0}^{H-1} \left({x}_{i}-\hat{x}_{i}\right)^2,
\end{eqnarray}
%where $\tilde{\mathbf{h}}$ is a constant vector with small entries. 
where $\eta$ is the weight of the sparsity penalty term;  $x_i$ and ${ \hat{x}_i}$ represent the input and reconstructed data, respectively, as shown in Fig.~\ref{Fig:AE_DCST}. 
 
\begin{algorithm}
 %\SetAlgoNoLine  
 \caption{Asymmetrical autoencoder with a sparsifying DCST layer for data compression}
 \label{algor}
  \KwIn{Training dataset $\mathbf{X}_{train}$, Testing dataset $\mathbf{X}_{test}$,
moving window length $L$, training epoch $E$, threshold $\xi$, batches amount $N_{\rm{batch}}$.}
  \KwOut{Reconstruction set $\mathbf{D}$,}
  $\mathbf{Data\ {preparation}}$: Process dataset using $L$;\\
  \tcc{Offline model training stage}
  $\mathbf{Step\  1}$: Initialize training epoch $e = 0$, PRD set $\mathbf{R}=\emptyset$, where $\emptyset$ is an empty set.\\
  \While{$\underline{e< E}$}
  {
        Compute the reconstruction output $\mathbf{z}$ in Eq.~(\ref{Eq:out});\\
        Compute the loss $\mathcal{L}$ in Eq.~(\ref{Eq:loss_p});\\
        Compute the PRD using Eq.~(\ref{Eq:prd});\\
        $\mathbf{R}\leftarrow{\mathbf{R}\cup{\rm{PRD}}};$\\
        \If {$\underline{{\rm{PRD}}\le \min\{\mathbf{R}\}}$} 
        {
            Save weights $\mathbf{W}$
        }                
        \If {$\underline{{\rm{the\ proportion\ of\ 0s\ in\ }} \mathbf{X}_{train}\le \xi}$} 
        {
            break
        }
        $e\leftarrow{e+1};$\\
  }
  \tcc{Online data transmission or storage stage}
  $\mathbf{Step\  2}$: Initialize batch index $i=0$, $\mathbf{D}=\emptyset$;\\  
    \While{$\underline{i < {N_{\rm{batch}}}}$}
  {
        $\mathbf{X}_{test}[i]\leftarrow\mathbf{X}_{test}[i].{\rm{reshape}}(-1)$\\
        Compute the coefficients $\mathbf h$ using Eq.~(\ref{Eq:HT});\\
        Convert $\mathbf h$ into bit streams $\mathbf d$ using hybrid coding;\\
        Reconstruct the signal $\mathbf z$ using Eq.~(\ref{Eq:out})\\
        $\mathbf{D}\leftarrow{\mathbf{D}\cup{\mathbf z}};$\\
        $i\leftarrow{i+1};$\\
  }
  return ${D}$; \\
  \end{algorithm}
\subsubsection{Training for asymmetrical autoencoder with a sparsifying DCST layer}

%The overall loss function $\mathcal{L}$ is 

The training algorithm 
%for the autoencoder with the DCST layer 
is summarised in Algorithm \ref{algor}. We save the weight set with the lowest percent root mean square difference (PRD) during the training process. The threshold vector $\mathbf{T} =[ T_1, T_2, ..., T_N]^T$ in the hard-thresholding block is initialized with large positive values, so it can set most of the DCST coefficients to zeros at the beginning of the training. This is also another factor leading to a sparse vector in the latent domain of the autoencoder. Whenever the proportion of zeros in the training set is less than a threshold $\xi$ during the training process, the training process is terminated. In this way, the coefficients with small values are removed (set as zero).
% \begin{figure*}[htbp]
% 	\centering
% 		\includegraphics[scale=.137]{figure/AE_DCST_T2.jpg}
% 	\caption{Training for asymmetrical autoencoder with DCST layer.}
% 	\label{Fig:AE_DCST_T}
% \end{figure*}

% As shown in Fig.~\ref{Fig:AE_DCST_T},
% the input vector $\mathbf{x}\in\mathbb{R}^{H}$. 
% %From 
% We considered two cases for $\sigma$ in Eq. (\ref{Eq:linear}) in our experimental studies.
% The output after the first linear layer is either
% \begin{eqnarray}\label{Eq:first_out}
% \mathbf{c}=\tanh{\left(\mathbf{W}^{(1)}\mathbf{x}+\mathbf{b}^{(1)}\right)}
% \end{eqnarray}
% or $\sigma$ is the identity function and $\mathbf{c}=\left(\mathbf{W}^{(1)}\mathbf{x}+\mathbf{b}^{(1)}\right)$.
% Then, the DCST layer calculates the DCST of the vector $\mathbf{c}$ and obtains the vector $\tilde{\mathbf{y}}$. 
% Then from Eq. (\ref{Eq:linear}), the output of 
% the second linear is 
% \begin{eqnarray}\label{Eq:out}
% \mathbf{z}=\mathbf{W}^{(2)}\tilde{\mathbf{y}}+\mathbf{b}^{(2)}
% \label{wholeoutput}
% \end{eqnarray}
%Suppose the output of hard-thresholding is $\mathbf{h}$.

\begin{figure*}[htbp]
	\centering
		\includegraphics[scale=.34]{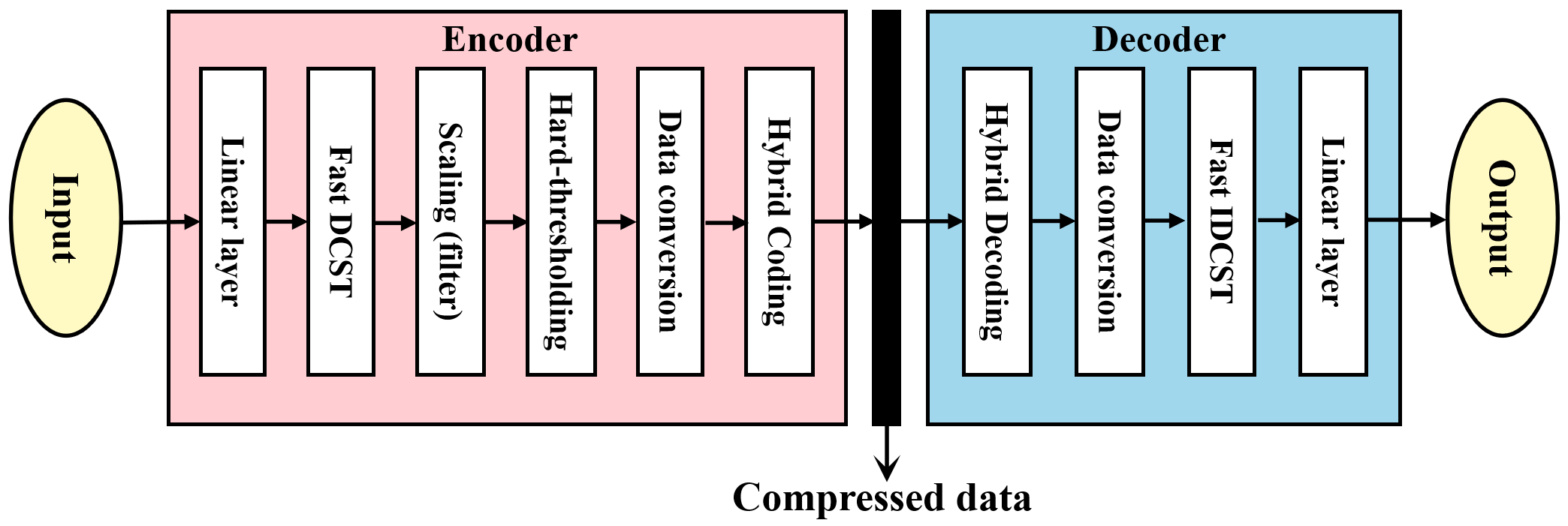}
	\caption{Data transmission or recording model}
	\label{Fig:online}
\end{figure*}

\subsection{Data Transmission or Storage}
\label{sec:transmission}
After the training process, the network is ready for data transmission or storage as shown in Fig. \ref{Fig:online}. At the encoder, the output of the hard-thresholding function in the DCST layer is converted into integers. Then, hybrid coding~\cite{akhter2010ecg} is employed to convert the data into bit streams as described in  Section ~\ref{sec:Hybrid coding}. The compressed signal is sent to the decoder or stored for later usage. The decoder applies hybrid decoding and data conversion to get the floating-type data. After that, the signals are reconstructed by the IDCST and the well-trained linear layer.

The DCST computation process is summarised in Fig.~\ref{Fig:DCST}. 
In general, a fully connected linear layer has a computation cost of $O(N^2)$ for an input vector of size $N$. Straightforward computation of DCT also requires $O(N^2)$ operations. However,
both the DCT and IDCT can be computed in $O(Nlog_2 N)$ operations using their fast algorithms for an input vector of size $N =2^p, \ p$ integer.
%During the training stage, we employ the product between the DCT weight matrix and the input tensor to perform the DCT in the DCST. 
Since the DCST is computed using the DCT, it is also an Order$(Nlog_2 N)$ algorithm as described in Algorithm~\ref{algorithm:Fast DCST}.
%For an input vector with a size of $N$, the numbers of real multiplications and real additions of the DCT are $N^2$ and $N^2-N$, respectively. Then, the DCST takes $\frac{4}{3}N^2+\frac{2}{3}$ real multiplications and $\frac{4}{3}N^2-2N+\frac{2}{3}$ real additions. During the transmission stage, to reduce the hardware cost in the sensor, we propose the fast DCST as shown in the Algorithm. 3. 
In summary, we first use a fast DCT algorithm~\cite{chen1977fast} to compute the DCT coefficients. The exact numbers of real multiplications and real additions for the fast DCT computations are $N\log_2 N-\frac{3}{2}N+4$ and $\frac{3}{2}N\log_2 N-\frac{3}{2}N+2$, respectively. After this step, DCT coefficients are grouped into small blocks and the IDCTs of the small blocks are computed to obtain the DCST coefficients as shown in Figure.~\ref{Fig:DCST}. As a result, the fast DCST requires $(2N+4)\log_2 N-5N+7$ real multiplications and $(3N+2)\log_2 N-6N+6$ real additions, respectively. 
In our implementation, the gearbox input block size is 80. The fully connected layer at the encoder has $N=64$ neurons. Therefore, the DCT and DCST sizes are $N=64$ and 
the fast DCST algorithm can be used at the edge to reduce the computational cost during data transmission and storage. 
% Moreover, during the training stage, we employ the product between the DCT weight matrix and the input tensor to perform the DCT in the DCST. 
% For an input vector with a size of $N$, the numbers of real multiplications and real additions of the DCT are $N^2$ and $N^2-N$, respectively. Then, the DCST takes $\frac{4}{3}N^2+\frac{2}{3}$ real multiplications and $\frac{4}{3}N^2-2N+\frac{2}{3}$ real additions. During the transmission stage, to reduce the hardware cost in the sensor, we propose the fast DCST as shown in the Algorithm. 3. In detail, we apply a fast DCT algorithm~\cite{chen1977fast} in the DCST. The numbers of real multiplications and real additions of the fast DCT are $N\log_2 N-\frac{3}{2}N+4$ and $\frac{3}{2}N\log_2 N-\frac{3}{2}N+2$, respectively. Next, the fast DCST takes $(2N+4)\log_2 N-5N+7$ real multiplications and $(3N+2)\log_2 N-6N+6$ real additions. Therefore, the computation cost of the DCST is decreased from Order$(N^2)$ to Order$(N\log_2 N)$ during the transmission stage. 
% We do not use this fast DCT algorithm during the training process because it is not supported in PyTorch officially. But it is suitable
% for implementation in hardware.
The online data transmission stage is summarised in Algorithm \ref{algor}.

\begin{algorithm}
 %\SetAlgoNoLine  %
 \caption{Fast DCST}
 \label{algorithm:Fast DCST}
  \KwIn{Input tensor ${\mathbf{c}} \in\mathbb{R}^{N}$}
  \KwOut{Output tensor ${\mathbf{X}} \in\mathbb{R}^{N}$}
        ${\mathbf{C}}={\textbf{Fast DCT}}(\mathbf{c})\in\mathbb{R}^{N}$; \\
        \For{$i=2;i \le \log_{2}N+1;i=i+1$}
        {
            $\beta=2^{i-2}$; \\
            $\lambda=\beta-1$;\\
            $\mathbf{X}[\beta:\lambda+\beta]={\textbf{Fast IDCT}}(\mathbf{C}[\beta:\lambda+\beta])\in\mathbb{R}^{N}$; \\
        }
        return ${\mathbf{X}}\in\mathbb{R}^{N}$; \\
\end{algorithm}

\subsubsection{Hybrid coding}
\label{sec:Hybrid coding}
%\subsubsection{Data conversion}

As pointed out above, some of the 
scaled and hard-thresholded DCST coefficients become zero. Other coefficients are floating-type parameters. To save memory, they are scaled by a predetermined value $\phi$ and rounded to $\theta$ decimal places. Then, data are multiplied by $10^{\theta}$ to be converted into integers. In this method, higher compression efficiency can be obtained with low precision loss. The data conversion is summarized as
\begin{eqnarray}\label{Eq:code}
\widehat{\mathbf{\mathcal{Q}}}=\text{Round}( 10^{\theta}\times \widetilde{\mathbf{\mathcal{Q}}}/\phi ) ,    
\end{eqnarray}
where $\widehat{\mathbf{\mathcal{Q}}}\in\mathbb{R}^N$ is the output of the data conversion operation.

This data is now ready for hybrid coding.
Run-length encoding (RLE)~\cite{akhter2010ecg} is a lossless encoding method, which can be used to exploit repeated bits or character sequences and replace them with their occurrence times. After the RLE, two lists are obtained. One contains the data with different values, and the other contains the number of occurrences of the corresponding data. Huffman coding~\cite{sharma2010compression} is a kind of variable length coding (VLC), which uses a shorter code for a letter with a higher frequency of occurrence and a longer code for a letter with a lower frequency.
\begin{figure}[htbp]
	\centering
		\includegraphics[scale=.33]{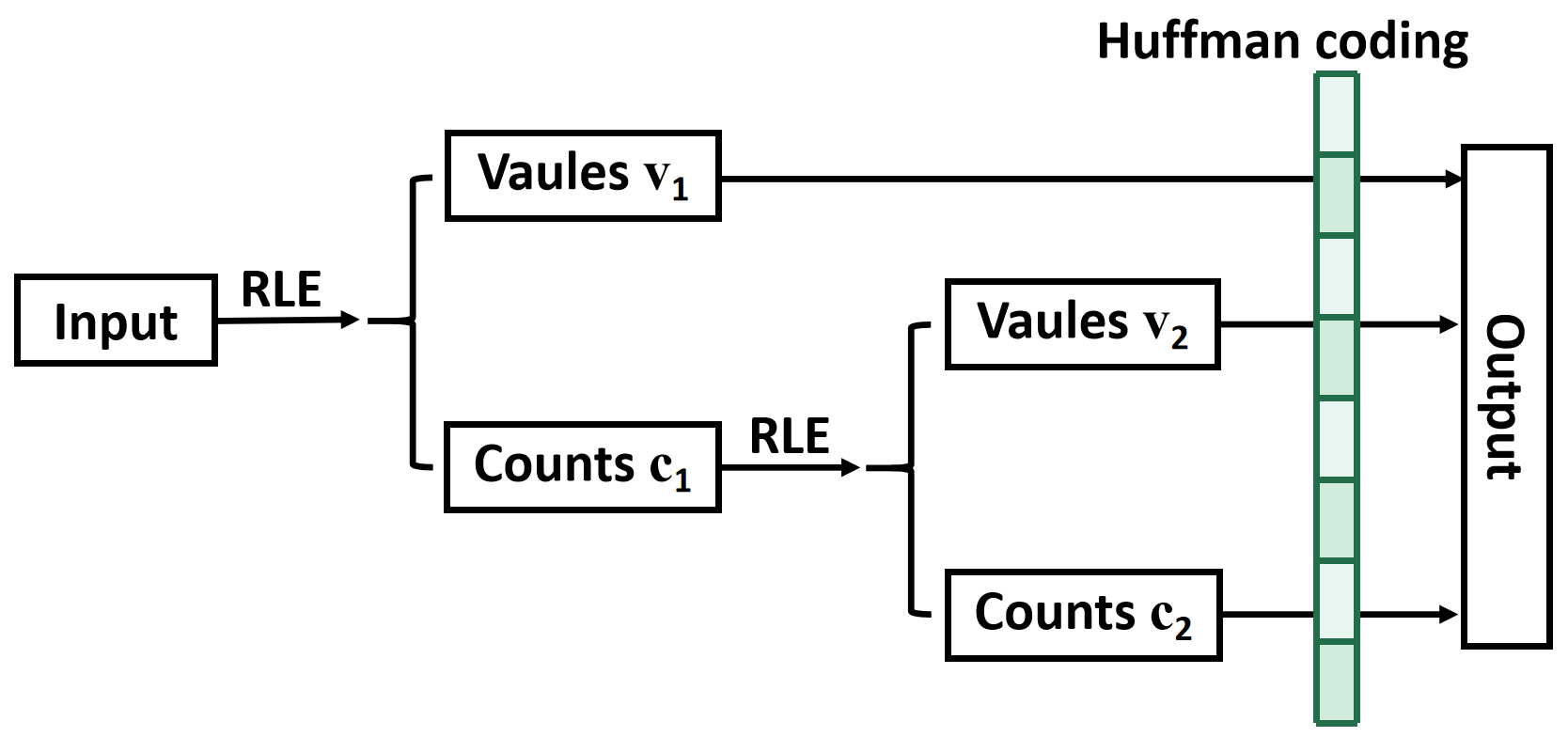}
	\caption{Block diagram of hybrid coding}
	\label{Fig:HC}
\end{figure}

In data compression tasks, the combination of the RLE and the Huffman coding can achieve a higher compression ratio. After the hard-thresholding layer, there are many zeros among the DCST coefficients. Therefore, the RLE is first used to remove zeros in the input as shown in Fig.~\ref{Fig:HC}. After that, the repeated "1"s appear in $\mathbf{c}_1$ because the DCST coefficients of the first half tend to be different. So, the RLE is used again to reduce the number of "1"s in $\mathbf{c}_1$. At the last step, Huffman coding is performed on $\mathbf{v}_1$, $\mathbf{v}_2$, and $\mathbf{c}_2$.

\section{Experimental verification}
\label{sec:Experiments}
In this paper, we use the following performance metrics to analyze the compression efficiency and data reconstruction accuracy of the proposed data compression algorithm:
\subsection{Performance metrics}
\label{sec:Metrics}

%Compression efficiency and data reconstruction accuracy are important criteria to measure the performance of compression models. 
The compression ratio (CR), percent root mean square difference (PRD), normalized percent root mean square difference (PRDN), root mean square (RMS), and quality score (QS)
are defined as follows:

\begin{itemize}

%\subsection{Compression ratio (CR)}
\item CR is expressed as the ratio of the original data size $L_r$ to the compressed data size $L_c$:
\begin{equation}
{\rm{CR}}=\frac{L_{r}}{L_{c}},
\end{equation}
CR shows the ability of the model to eliminate redundant data. The higher the CR is, the better the model is.

%\subsection{Root mean square error (RMS)}
\item RMS indicates the difference between the original data and the reconstructed data:
\begin{equation}
{\rm{RMS}}=\sqrt{\frac{1}{k}\sum_{i=0}^{K-1}\left(d_i^o-d_i^r\right)^2}\times 100,
\end{equation}
where $K$ is the data size; $d_i^o$ is the element of original data; and $d_i^r$ is the element of reconstructed data. The lower the RMS is, the better the model is.

%\subsection{Percent root mean square difference (PRD)}
\item PRD also measures the reconstruction distortion, which indicates the quality of reconstructed data: 
\begin{equation}
{\rm{PRD}}=\sqrt{\frac{\sum_{i=1}^{K}\left(d_i^o-d_i^r\right)^2}{\sum_{i=1}^{K}\left(d_i^o\right)^2}}\times 100,
\label{Eq:prd}
\end{equation}
 The lower the PRD is, the better the model is. The PRD is preferred in data compression applications because it normalizes the RMS using the energy of the original input signal.
 
%\subsection{Normalized percent root mean square difference (PRDN)}
\item PRDN is the normalized version of PRD. It is expressed as follows:
\begin{equation}
{\rm{PRDN}}=\sqrt{\frac{\sum_{i=1}^{K}\left(d_i^o-d_i^r\right)^2}{\sum_{i=1}^{K}\left(d_i^o-\bar{d}\right)^2}}\times 100,
\end{equation}
where $\bar{d}$ is the average of the original data. We include PRND because PRD is sensitive to original data. The lower the PRDN is, the better the model is.

%\subsection{Quality score (QS)}
\item QS indicates the tradeoff between the compression ratio and the reconstruction accuracy:
\begin{equation}
\rm{QS}=\frac{CR}{PRD},
\end{equation}
Usually, the reconstruction accuracy declines when the data compression ratio increases. QS involves both the compression ratio and the accuracy of the reconstructed signal. The higher QS is, the better the model is.
\end{itemize}

%\subsection{Open gearbox data}
The data compression on publicly available gearbox datasets has not been studied in detail in recent literature. To demonstrate the performance of the compression algorithm, we use two open gearbox datasets: the Southeast University (SEU) gearbox dataset~\cite{seu,shao2018highly}, and the University of Connecticut (UoC) gear fault dataset~\cite{uoc} in this work:
\begin{itemize}
%\subsubsection{SEU Gearbox Dataset}
\item SEU gearbox dataset is collected from the Drivetrain Dynamic Simulator and it contains two kinds of working states with rotating speeds: 20Hz and 30Hz. There are five kinds of gearbox faults in each working state including chipped tooth, missing tooth, root fault, surface fault, and healthy working state. Therefore, it is divided into a total of ten data files. Each file contains eight columns of vibration signals. The signals of columns 2, 3, and 4 are all effective~\cite{seu}. Hence, we use the third column of each data file. It indicates the vibration signal of the planetary gearbox in the Z-orientation. These data were sampled at 2000Hz. 
%As tabulated in Table~\ref{table: SEUF}
% \begin{table}[htbp]
% \centering
% \caption{SEU gearbox datasets description.}\label{table: SEUF}
% \begin{tabular}{ccc}
% \toprule
% Fault Type    & Rotating Speed & Load Configuration \\ \midrule
% Chipped Tooth & 20 Hz          & 0 V                \\
% Chipped Tooth & 30 Hz          & 2 V                \\
% Health Gear   & 20 Hz          & 0 V                \\ 
% Health Gear   & 30 Hz          & 2 V                \\
% Missing Tooth & 20 Hz          & 0 V                \\
% Missing Tooth & 30 Hz          & 2 V                \\
% Root Fault    & 20 Hz          & 0 V                \\
% Root Fault    & 30 Hz          & 2 V                \\
% Surface Fault & 20 Hz          & 0 V                \\
% Surface Fault & 30 Hz          & 2 V                \\
% \bottomrule
% \end{tabular}
% \end{table}
%\subsubsection{UoC Gear Dataset}
\item UoC gear fault datasets were sampled at 20 kHz. It records nine different gear fault types including missing tooth, root crack, healthy state, spalling fault, and chipping tip with 5 different levels of
severity~\cite{zhao2020deep}. Since the last five faults belong to the chip, we use chipping tips with two typical severity levels in this experiment: chip1a and chip2a.
\end{itemize}

\subsection{Dataset Preprocessing}
SEU and UoC datasets are in the floating-type format. However, the format of the analog-to-digital converter output is usually expressed by an integer. Therefore, we multiply all floating-type numbers by $10^6$ and round them. This operation is conducive to making the data available in a fair manner.
%Then we use an 80-length window to split the data. 
We select the top 20\% of the data as the training set and the remaining 80\% as the testing set. Additionally, we divide all data by the maximum value of the training set to realize normalization.

\subsection{Comparison With Other Methods}
\label{sec:Comparison}

To overcome the limitations of low reconstruction accuracy in transform-based methods and over-smoothing in autoencoders, we propose an autoencoder with the DCST layer, which has trainable parameters including the thresholds and the scaling parameters that emphasize or deemphasize frequency bands. These parameters are learned from data and the only hyperparameter is the block size. Since the DCST has a fixed frequency domain structure covering all the frequency components of the input it also prevents "low-pass filtering" performed by standard autoencoders. 
 In gearbox data compression trainable hard-thresholding units not only remove the noise in the data but also improve the data compression efficiency. 
A single DCT, DCST, fully connected or convolutional layer cannot be trained together with a soft-thresholding or hard-thresholding nonlinearity. However, by adding the fixed DCST after a fully connected layer we can not only train the thresholds but also train the fully connected layer which adapts the data and improves the data compaction capability 
of DCST.

The proposed compression scheme is developed in Python code on a PC with Intel 4.7 GHz CPU, and 16 GB RAM. As pointed above we divide the gearbox signals into blocks of size $H=80$. To train the neural networks, we use the loss function defined in Eq.~(\ref{Eq:loss_p}) and use the AdamW optimizer~\cite{loshchilov2017decoupled}.  $\eta$ is selected as 100. ${\gamma}$ is 0.0001.
The batch size and the learning rate of each experiment are set to 16 and 0.001, respectively. 
% As described in section~\ref{sec:Introduction}, some classical or new deep autoencoders are not suitable for gearbox sensor data compression. 
Compression models are expected to have a low computational complexity on the encoder side, making them suitable for implementation in low-cost sensors.
Therefore, the multi-layer autoencoder~\cite{tan2008performance}, sparse autoencoder~\cite{ng2011sparse}, standard DCST~\cite{978},
% autoencoder with a DCT layer~\cite{pan2022real}, 
hybrid quantum-classical Hadamard transform perceptron (HQHTP)~\cite{pan2023hybrid}, variational autoencoder~\cite{oliveira2023early}
and JSNet~\cite{chen2020jsnet} are selected as comparison models. Huffman coding is used in the latent space to encode the compressed signals.
% Moreover, to compare DCST with DCT, a reference experiment is set up. In this experiment, DCT is used to replace DCST in the asymmetrical autoencoder.

The structure of the multi-layer autoencoder is shown in Fig.~\ref{Fig:AE}. In the experiment, there are four linear layers in the multi-layer autoencoder and sparse autoencoder. Tensors in the autoencoders go as $\mathbb{R}^{80}\rightarrow\mathbb{R}^{64}\rightarrow\mathbb{R}^{48}\rightarrow\mathbb{R}^{64}\rightarrow\mathbb{R}^{80}$. The waveform of the reconstructed rotating machine signal is close to that of the original signal when the PRD is around 24~\cite{guo2013novel}. Therefore, to ensure high reconstruction accuracy of the model, $B$ is set as $48$, which makes the PRD of auto-encoders around 24 in gearbox data compression. 
To make a fair comparison with other types of autoencoders, we only keep 38 DCST coefficients when the threshold $\xi$ is 0.4 and 32 DCST coefficients when $\xi$ is 0.5. 
% We choose 32 and 38 because the coefficients that are set to 0s still need more bits to transmit. 
% Then, $\xi$ is increased to eliminate the negative impact of removed coefficients on the compression ratio. 
In the standard DCST algorithm, the top sixty percent of DCST coefficients are retained. Other coefficients are set as zero.
% Moreover, when $\xi=0.4$ and $0.5$, each element in the threshold vector $\mathbf{T}$ is initialized as 0.25 and 0.35 respectively.
Similarly, suitable threshold vectors are selected to make other methods
have a similar number of training iterations for a fair comparison.

\begin{table}
\caption{Compression experiment on the SEU dataset.}
\label{tab: Compression experiment on SEU dataset}
\begin{tabular}{c c c c c c c}
\hline
\multirow{2}{*}{\makecell{\textbf{Algorithm}}}&\textbf{Metrics}&\textbf{20$\rightarrow$20}  \\ \cmidrule[0.75pt]{2-7}
        \multirow{2}{*}{ } &\textbf{CR}
        &\textbf{PRD}&\textbf{PRDN}&\textbf{RMS}& \textbf{QS}& \textbf{N$_{\mathbf{coe}}$}  \\ 
\hline

           Multi-layer AE &8.93   & 22.95    &23.05    &2.49    &0.39 &48\\
            Sparse AE &9.76    &22.95    &23.05    &2.49    &0.43 &48\\
            {Variational AE} &9.30 & 22.90 & 23.00 & 2.48 & 0.41& 48\\
            Standard DCST &9.16 & 26.43 & 26.55 & 2.84 & 0.35& 38\\
            {JSNet} &8.49 & 28.05 & 28.17 & 3.01 & 0.31 &-\\
            {HQHTP} & 9.27  &22.24   &22.34    &2.41   &0.42 & -\\
            AE-DCT($\xi=0.4$) &9.68 & 19.32 & 19.40 & 2.10 & 0.50 &42\\
            AE-DCT($\xi=0.5$) &10.06 & 22.92 & 23.02 & 2.49 & 0.44&34\\    
            \textbf{AE-DCST($\xi=0.4$)} &9.68 & \textbf{18.95} & \textbf{19.03} & \textbf{2.05} & \textbf{0.51} &42\\
            \textbf{AE-DCST($\xi=0.5$)} &\textbf{10.11} & 22.34 & 22.44 & 2.42 & 0.45 &35\\  \hline     
           % \hline     
\end{tabular}
\begin{tablenotes}
        \footnotesize
\item Note: \textbf{AE-DCST} is the proposed method. AE-DCT represents the asymmetrical autoencoder where the DCST is replaced with the DCT.
      \end{tablenotes}
\end{table}

In the compression experiment on the SEU dataset, we use the data with a rotating speed of 20Hz to train and test models.
We split the SEU dataset into 2,621 training samples and 10,486 testing samples. Additionally, we choose $\phi=3.5$ and $\theta=3$ in Eq.~(\ref{Eq:code}). 
Table~\ref{tab: Compression experiment on SEU dataset} provides the average compression results for SEU datasets. The detailed results on five gearbox fault types are put in the supplementary material. 
% Compared with traditional autoencoders, although the autoencoder with a DCT layer obtains a better average CR, its average PRD is higher. 
Among the familiar gearbox data lossy compression methods, the average PRD of the AE-DCST($\xi=0.4$) is only 18.95 while the compression ratio is about 9.68. Therefore, the AE-DCST($\xi=0.4$) can achieve a balance between compression efficiency and reconstruction accuracy. When DCT is used to replace DCST, results from Table~\ref{tab: Compression experiment on SEU dataset} show that there is a drop in performance. The CR and PRD of the AE-DCT($\xi=0.5$) scheme are 10.06 and 22.92 respectively, but these of the AE-DCST($\xi=0.5$) are up to 10.11 and 22.34. Even though the computational complexity of DCST is larger than that of DCT, it reduces transmitting bits and improves reconstruction accuracy. It shows that scaling in the DCST domain can achieve better results than in the DCT domain, which indicates convolution can extract more important features than scaling operation. DCST uses DCT to divide the data into subbands similar to the audio compression methods therefore it is more suitable for streaming sensor data applications than a straightforward application of the DCT onto the gearbox data.
Compared with standard DCST, the proposed model has better performance because the linear layer and trainable scaling parameters in AE-DCST  enhance the ability to extract features.
In addition, by changing the value of $\xi$ in the transform domain, we can reconstruct the original signal at various quality levels.

To further demonstrate the robustness of the proposed model, the trained model is tested on the datasets labeled as other working conditions: a rotating speed of 30Hz. The results in Table~\ref{tab: Transfering learning on the SEU dataset} show that the proposed DCST-based model has the best PRD, PRDN, RMS, and QS among the lossy compression methods. Therefore, the proposed model has a better generalization ability than other methods. 
%This improvement also proves the DCST layer's effectiveness.
\begin{longtable}{c c c c c c c}
\caption{Transfering learning on the SEU dataset.}
\label{tab: Transfering learning on the SEU dataset}\\
\hline
\multirow{2}{*}{\makecell{\textbf{Algorithm}}}&\textbf{Metrics}&\textbf{20$\rightarrow$30}  \\ \cmidrule[0.75pt]{2-7}
        \multirow{2}{*}{ } &\textbf{CR}
        &\textbf{PRD}&\textbf{PRDN}&\textbf{RMS}& \textbf{QS}& \textbf{N$_{\mathbf{coe}}$}  \\ 
\hline
\endfirsthead
\multicolumn{7}{c}%
{\tablename\ \thetable\ -- \textit{Continued from previous page}} \\
\hline
%  \multirow{2}{\textbf{Algorithm}} & \textbf{CR} & \textbf{PRD}& \textbf{PRDN} & \textbf{RMS} & \textbf{QS} & \textbf{N$_{\mathbf{coe}}$}\\
% \hline

\multirow{2}{*}{\textbf{Algorithm}}&\textbf{Metrics}  \\ \cmidrule[0.75pt]{2-7}
        \multirow{2}{*}{ } &\textbf{CR}
        &\textbf{PRD}&\textbf{PRDN}&\textbf{RMS}& \textbf{QS} & \textbf{N$_{\mathbf{coe}}$} \\ 
\hline

\endhead
\hline \multicolumn{7}{r}{\textit{Continued on next page}} \\
\endfoot
\hline
\endlastfoot
            Multi-layer AE &8.86 & 21.75 & 21.77 & 2.90 & 0.41 &48\\
            Sparse AE &9.56    &21.71    &21.73    &2.89    &0.44 &48
          \\
          {Variational AE} &9.15 & 21.67 & 21.69 & 2.89 & 0.42& 48\\
            Standard DCST&9.38 & 21.81 & 21.83 & 2.89 & 0.43& 38\\
           {JSNet} &8.42 & 22.75 & 22.76 & 3.01 & 0.37&-
          \\
           {HQHTP} &9.39 & 20.87 & 20.89 & 2.78 & 0.46& -\\
            AE-DCT($\xi=0.4$) & 9.58 & 17.45 & 17.46 & 2.33 & 0.55 &45

          \\
            AE-DCT($\xi=0.5$) &10.15 & 20.93 & 20.94 & 2.80 & 0.49 &38
          \\    
            \textbf{AE-DCST($\xi=0.4$)} &9.56 & \textbf{16.99} & \textbf{17.01} & \textbf{2.27} & \textbf{0.57} &45
          \\
            \textbf{AE-DCST($\xi=0.5$)} & \textbf{10.15} & 20.34 & 20.36 & 2.71 & 0.50&38
          \\        
            %\hline
\end{longtable}

The UoC dataset is split into 936 training samples and 3,744 testing samples. During data transmission, we set $\phi=4.5$ and $\theta=3$ in Eq.~(\ref{Eq:code}).  The average compression results for the UoC datasets are summarized in Table \ref{tab: Compression experiment on Uoc datasets}. 
The detailed results on different gearbox fault types are attached to the supplementary material. 
Compared with the sparse autoencoder which outperforms the standard autoencoder and JSNet, the AE-DCST($\xi=0.4$) reduces the average PRD from 13.78 to 10.95 (20.54\%), RMS from 2.86 to 2.28 (20.28\%). Additionally, it increases the CR from 9.33 to 9.63 (3.22\%) and QS from 0.72 to 0.90 (25\%), respectively. 
% Moreover, as shown in Table \ref{table: np}, the AE-DCST has lower MACs than other autoencoders. 
The sparse autoencoder and the AE-DCST model utilize the same sparse penalty function. When the DCST layer is introduced to replace the middle two linear layers in a sparse autoencoder, the performance of the model has been significantly improved.
It indicates the DCST layer has a better compression efficiency than the linear layer. 
Compared with the variational AE, the AE-DCST achieves a higher CR and a lower PRD. 
Although the variational AE incorporates Gaussian processes to extract more features, it still uses a traditional autoencoder structure. Its imperfect encoder and decoder module lead to feature loss, which increases reconstruction errors. Moreover, AEs work as low-pass filters. High-frequency components are removed during the reconstruction process. In contrast, although the DCST layer removes the small entries in the frequency domain, it still retains the large entries in the high-frequency bands. Therefore, the AE-DCST captures more important features than other autoencoder-based models.
% Additionally, It is experimentally shown in section 4.5 that the DCST layer in the AE-DCST can retain more important features than the linear layer in other autoencoder-based methods during the compression process. 
Besides, it is observed that the standard DCST, JSNet and HQHTP lag behind the AE-DCST model in terms of both CR and PRD. The reason is that the AE-DCST employs fully connected layers to enhance the feature extraction ability of the encoder module and the reconstruction ability of the decoder module. 
% Furthermore, DCST uses DCT to divide the data into subbands similar to the audio compression methods therefore it is more suitable for streaming gearbox sensor data compression.

% Besides, the numbers of trainable parameters in the sparse autoencoder and proposed model are  16,640 and 10,512 respectively.
% %This improvement shows the effectiveness of the DCST layer. 
% It proves that the AE-DCST model can achieve higher accuracy with fewer trainable parameters.

\begin{longtable}{c c c c c c c}
\caption{Compression experiment on the UoC datasets}\label{tab: Compression experiment on Uoc datasets}\\
\hline
\multirow{2}{*}{\makecell{\textbf{Algorithm}}}&\textbf{Metrics}  \\ \cmidrule[0.75pt]{2-7}
        \multirow{2}{*}{ } &\textbf{CR}
        &\textbf{PRD}&\textbf{PRDN}&\textbf{RMS}& \textbf{QS}& \textbf{N$_{\mathbf{coe}}$}  \\ 
\hline
\endfirsthead
\multicolumn{7}{c}%
{\tablename\ \thetable\ -- \textit{Continued from previous page}} \\
\hline
%  \multirow{2}{\textbf{Algorithm}} & \textbf{CR} & \textbf{PRD}& \textbf{PRDN} & \textbf{RMS} & \textbf{QS} & \textbf{N$_{\mathbf{coe}}$}\\
% \hline

\multirow{2}{*}{\textbf{Algorithm}}&\textbf{Metrics}  \\ \cmidrule[0.75pt]{2-7}
        \multirow{2}{*}{ } &\textbf{CR}
        &\textbf{PRD}&\textbf{PRDN}&\textbf{RMS}& \textbf{QS} & \textbf{N$_{\mathbf{coe}}$} \\ 
\hline

\endhead
\hline \multicolumn{7}{r}{\textit{Continued on next page}} \\
\endfoot
\hline
\endlastfoot
            Multi-layer AE &8.88 & 13.86 & 13.87 & 2.88 & 0.68 & 48\\
            Sparse AE &9.33 & 13.78 & 13.78 & 2.86 & 0.72 & 48\\
            {Variational AE} &8.97 & 13.58 & 13.58 & 2.82 & 0.71& 48\\
           Standard DCST &9.37 & 13.60 & 13.60 & 2.83 & 0.70 & 38\\
           {JSNet} &8.20 & 14.50 & 14.50 & 3.01 & 0.57 &-\\
            {HQHTP} &8.93 & 12.62 & 12.62 & 2.61 & 0.72& -\\
            AE-DCT($\xi=0.4$) &9.62 & 11.30 & 11.31 & 2.36 & 0.88 & 44\\
            AE-DCT($\xi=0.5$) &10.01 & 13.70 & 13.70 & 2.86 & 0.75 & 37\\
             \textbf{AE-DCST($\xi=0.4$)} &9.63 & \textbf{10.95} & \textbf{10.95} & \textbf{2.28} & \textbf{0.90} & 44\\
            \textbf{AE-DCST($\xi=0.5$)} &\textbf{10.03} & 13.49 & 13.49 & 2.81 & 0.76 & 37\\

\end{longtable}

As tabulated in Tables \ref{tab: Compression experiment on SEU dataset},
\ref{tab: Transfering learning on the SEU dataset},
and \ref{tab: Compression experiment on Uoc datasets}, $N_{coe}$ represents the average number of latent space coefficients retained in a single block of the testing set. Additionally, ``-" indicates that no coefficients are removed in the latent space. In comparison to other lossy compression methods, the proposed asymmetrical autoencoders use fewer coefficients and achieve a higher reconstruction accuracy. This is due to the fact that DCT or DCST-based methods have a good energy concentration property. Therefore, the asymmetrical autoencoders only need a small number of coefficients to reconstruct the signals with high accuracy.

\begin{figure*}[!ht]
	\centering
		\includegraphics[scale=.12]{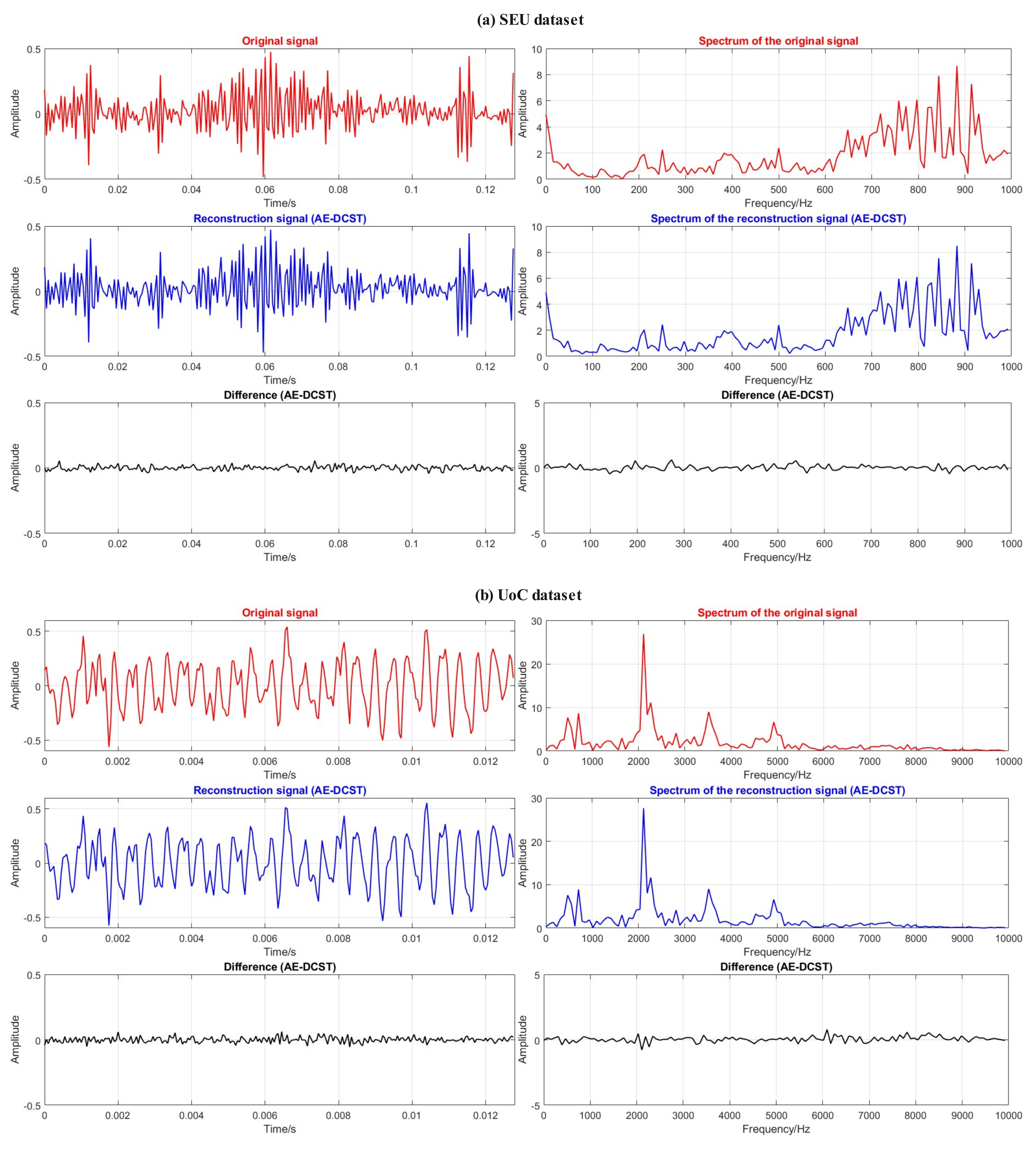}
	\caption{Comparison between the original data and the reconstructed data in the time domain (left) and the frequency domain (right) on the SEU dataset and UoC dataset.}
	\label{SEUF}
\end{figure*}

For visual assessment of the performance of the proposed data compression model, we plot the reconstruction signals in the time domain and frequency domain as shown in Fig.~\ref{SEUF}. Through visual inspection, the differences between the signals reconstructed by the AE-DCST and the original signals are limited to a small range in the time domain and frequency domain.  
Therefore, the reconstructed signals are very similar to the original signals. Additionally, it is observed that the AE-DCST can not only retain the main components in the low-frequency bands but also retain the important information in the high-frequency bands. Hence, the DCST layer has a good high-frequency computing capability to capture the important features in the frequency domain. Besides, the reconstructed signals of different methods are compared in Figs. 1-4 in the supplementary material. It is noticed that the signals reconstructed by the AE-DCST are closer to the original signals in the time domain in comparison to other compression methods.

% \begin{figure*}[!ht]
% 	\centering
% 		\includegraphics[scale=.35]{UoC.png}
% 	\caption{Comparison of the time domain between the original data (red) and the reconstructed data (blue).}
% 	\label{UocF}
% \end{figure*}

{The comparison of the number of trainable parameters and Multiply Accumulates (MACs) of the encoder part is presented in Table \ref{table: np}. When the input block size is 80, the AE-DCST has fewer training parameters and lower computational complexity in comparison to sparse AE, multi-layer AE and variational AE. This is because the DCST layer has a computation cost of $O(Nlog_2 N)$ for an input vector of size $N$. Therefore, it has a lower computational cost than the fully connected layers in other autoencoder-based methods. 
Additionally, since the standard DCST and JSNet do not include linear layers in the encoder, their computation costs are lower than AE-DCST. 
However, the standard DCST and JSNet have a lower compression efficiency than the AE-DCST as shown in Table~\ref{tab: Compression experiment on SEU dataset}. Moreover, AE-DCST and AE-DCT have a similar computational complexity and number of training parameters because they share a similar structure but AE-DCST has a higher compression ratio and reconstruction accuracy as shown in Table~\ref{tab: Compression experiment on Uoc datasets}.
\begin{table}[htbp]
\centering
\caption{The comparison of the number of trainable parameters and MACs.}\label{table: np}
%\begin{threeparttable}
\begin{tabular}{cccc}
\toprule
\multirow{2}{*}{\makecell{\textbf{Algorithm}}}   & \textbf{Input Block} & \textbf{Trainable} &\textbf{MACs}  \\ 
{}& \textbf{Size}& \textbf{Parameters}& \textbf{(Encoder side)}\\
\midrule
JSNet    & 1$\times$256         & 8,576       &16,640     \\
Standard DCST & 1$\times$64        & 0       &479        \\
Sparse AE & 1$\times$80        & 16,640        &8,192      \\
Multi-layer AE & 1$\times$80          & 16,640       &8,192        \\
Variational AE  &1$\times$80         & 11,696          &7,728   \\
AE-DCT & 1$\times$80         & 10,512             &5,476\\
\textbf{AE-DCST}   & 1$\times$80          &10,512       &5,663    \\
\bottomrule
\end{tabular}
%\end{threeparttable}
\end{table}

\subsection{Ablation study}
In this section, the ablation study is carried out to verify the function of each module in the proposed model. In the compression task, CR and PRD are two important performance metrics to measure compression performance. Hence, scaling and integer parameters are adjusted to optimize CR and PRD in each method, which can truly prove the effectiveness of different modules. $\xi$ is set as 0.5. Table \ref{SEU2} provides the average results on SEU dataset and UoC dataset. Detailed results are put in the supplementary material. 

\subsubsection{Ablation study on the scaling} 
As shown in Table \ref{SEU2}, when the scaling module is not present in the DCST layer, the performance degrades 
on both datasets, as a lower CR, a higher PRD, and a lower QS are obtained. It is because the scaling operation in the DCST domain is similar to the convolution operation in the DCT domain. This helps extract important features for the gearbox data compression. Furthermore, initialization with a sample quantization matrix contributes to the improvement of the compression ratio. Therefore, the scaling operation is mainly implemented to obtain more effective information and decrease the values of DCST coefficients. 

\subsubsection{Ablation study on the sparsity penalty}
As is shown in Table~\ref{SEU2}, when there is no sparsity penalty in the network,
the average CR is reduced from 10.11 to 9.86 (2.47\%) on the SEU dataset and from 10.03 to 9.99 (0.4\%) on the UoC dataset. These reductions indicate more bits are required to transmit signals. The average PRD increases from 22.34 to 27.22 (21.84\%) on the SEU dataset and from 13.49 to 16.75 (24.17\%) on the UoC dataset. Hence, adding a sparsity penalty is beneficial to ensure the sparsity of neurons while improving the accuracy of reconstruction. 

\subsubsection{Ablation study on the activation function}
As shown in Table~\ref{SEU2}, the reconstruction accuracy decreases when the activation function is removed from the proposed model. The reason for this is that models are limited to representing some linear relationships without any activation functions. However, gearbox data contains a large number of nonlinear components due to the complex interplay of various mechanical elements and external factors. Thus it is necessary to use an appropriate activation function in the model. 
\begin{longtable}{ccccccc}
\caption{Ablation experiment on SEU dataset and UoC dataset}\label{SEU2}\\
\hline
\multirow{2}{*}{\makecell{\textbf{Dataset}}}&\multirow{2}{*}{\makecell{\textbf{Algorithm}}}&\textbf{Metrics}  \\ \cmidrule[0.75pt]{3-7}
        \multirow{2}{*}{ } &\multirow{2}{*}{ }&\textbf{CR}
        &\textbf{PRD}&\textbf{PRDN}&\textbf{RMS}& \textbf{QS}  \\ 
\hline
\endfirsthead
\multicolumn{7}{c}%
{\tablename\ \thetable\ -- \textit{Continued from previous page}} \\
\hline
%  \multirow{2}{\textbf{Algorithm}} & \textbf{CR} & \textbf{PRD}& \textbf{PRDN} & \textbf{RMS} & \textbf{QS} & \textbf{N$_{\mathbf{coe}}$}\\
% \hline

\multirow{2}{*}{\makecell{\textbf{Dataset}}}&\multirow{2}{*}{\makecell{\textbf{Algorithm}}}&\textbf{Metrics}  \\ \cmidrule[0.75pt]{3-7}
        \multirow{2}{*}{ } &\multirow{2}{*}{ }&\textbf{CR}
        &\textbf{PRD}&\textbf{PRDN}&\textbf{RMS}& \textbf{QS}  \\ 
\hline

\endhead
\hline \multicolumn{7}{r}{\textit{Continued on next page}} \\
\endfoot
\hline
\endlastfoot
          &  No activation&9.96 & 23.35 & 23.46 & 2.53 & 0.43\\
           {SEU}&  No scaling &10.10 & 22.90 & 23.00 & 2.47 & 0.44\\
          {Dataset}&  No penalty&9.86 & 27.22 & 27.33 & 2.96 & 0.36\\
          &  \textbf{AE-DCST} &\textbf{10.11} & \textbf{22.34} & \textbf{22.44} & \textbf{2.42} & \textbf{0.45} \\\hline
          &  No activation&9.82 & 13.83 & 13.84 & 2.88 & 0.75\\
          {UoC}&  No scaling &9.94 & 15.16 & 15.16 & 3.16 & 0.66\\
          {Dataset}&  No penalty&9.99 & 16.75 & 16.76 & 3.48 & 0.62\\
          &  \textbf{AE-DCST} &\textbf{10.03} & \textbf{13.49} & \textbf{13.49} & \textbf{2.81} & \textbf{0.76}\\
          %\hline

\end{longtable}

\begin{figure*}[!ht]
	\centering
		\includegraphics[scale=.11]{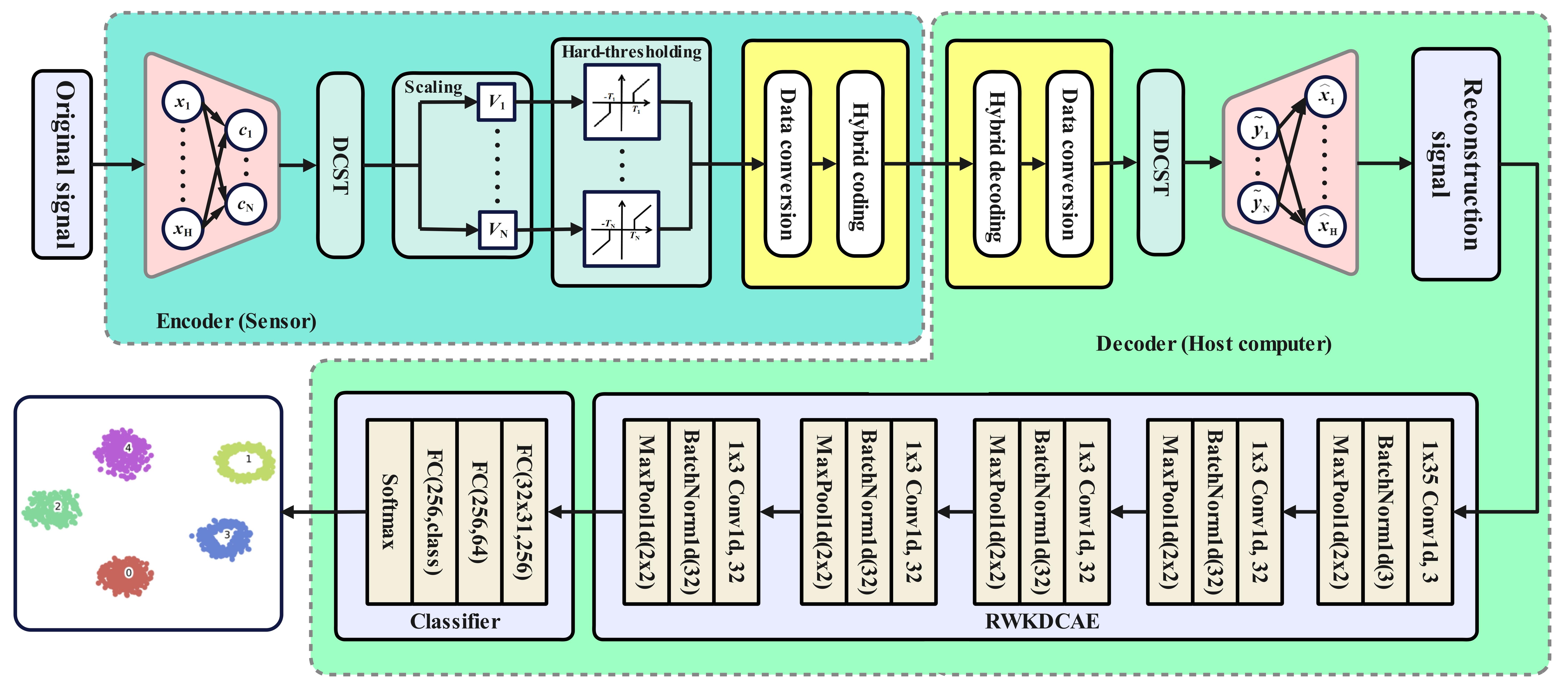}
	\caption{Fault diagnosis model consisting of the data compression/decompression and RWKDCAE network.}
	\label{Fig: Fault diagnosis model}
\end{figure*}
\subsection{Fault diagnosis experiment} 
The gearbox data compression model is designed for the remote fault detection system. In this section, fault diagnosis experiments are used to verify the validity of the data compression model. 
As is shown in Fig.~\ref{Fig: Fault diagnosis model}, we first apply the pre-trained encoder of the AE-DCST to perform gearbox data compression. Next, the compressed data are encoded into bitstreams for transmission. After that, the pre-trained decoder module of the AE-DCST reconstructs the original signal. 
% The reconstructed signal of the SEU dataset includes five types of faults at the rotating speed of 20Hz. Additionally, in the reconstructed signal of the UoC dataset, six main different gear conditions mentioned in the compression experiment are tested. 
Then, we select the top 70\% of the reconstructed data as the training set and the remaining 30\% as the testing set. Subsequently, we utilize the residual wide-kernel deep convolutional autoencoder (RWKDCAE) \cite{yang2021residual} as the fault diagnosis model. The RWKDCAE is designed based on convolutional layers, batch normalization layers, and pooling layers. The input to RWKDCAE is divided into blocks. Each block contains 1024 sample points. 
Since the authors in \cite{yang2021residual} also perform fault diagnosis using RWKDCAE on the SEU dataset, we follow the same training and testing procedures as described in their work.

\begin{figure*}[!ht]
	\centering
		\includegraphics[scale=.16]{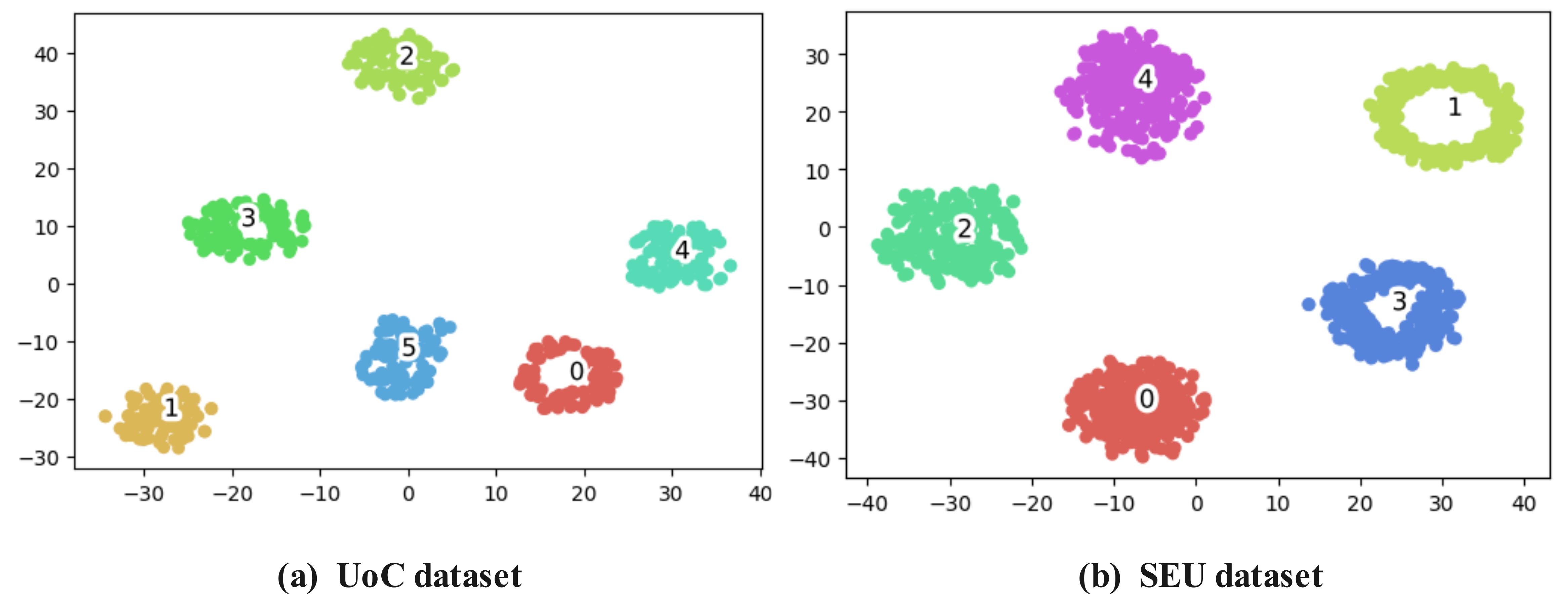}
	\caption{Feature visualization of RWKDCAE model on two datasets.}
	\label{feature}
\end{figure*}
\begin{longtable}{c c c }
\caption{{Fault diagnosis experiment on the SEU and UoC datasets.}}\label{tab: Fault diagnosis experiment}\\
\hline
\multirow{2}{*}{\makecell{\textbf{Algorithm}}}&\textbf{Accuracy}  \\ \cmidrule[0.75pt]{2-3}
        \multirow{2}{*}{ } &\textbf{SEU dataset} &\textbf{UoC dataset} \\ 
\hline
\endfirsthead
\multicolumn{3}{c}%
{\tablename\ \thetable\ -- \textit{Continued from previous page}} \\
\hline
%  \multirow{2}{\textbf{Algorithm}} & \textbf{CR} & \textbf{PRD}& \textbf{PRDN} & \textbf{RMS} & \textbf{QS} & \textbf{N$_{\mathbf{coe}}$}\\
% \hline

\multirow{2}{*}{\textbf{Algorithm}}&\textbf{Accuracy}  \\ \cmidrule[0.75pt]{2-3}
        \multirow{2}{*}{ } &\textbf{SEU dataset}
        &\textbf{UoC dataset}\\ 
\hline

\endhead
\hline \multicolumn{3}{r}{\textit{Continued on next page}} \\
\endfoot
\hline
\endlastfoot
            Multi-layer AE+RWKDCAE &100.00\% & 99.24\% \\
            Sparse AE+RWKDCAE &99.92\% & 99.05\% \\
           {Variational AE+RWKDCAE} &99.92\% & 99.81\% \\
           Standard DCST+RWKDCAE &99.02\% & 99.62\%\\
            {JSNet+RWKDCAE} &100.00\% & 99.81\% \\
           {HQHTP+RWKDCAE} &100.00\% & 99.62\% \\
            AE-DCT+RWKDCAE &100.00\% & 100.00\% \\
             \textbf{AE-DCST+RWKDCAE} &\textbf{100.00}\% & \textbf{100.00\%} \\

\end{longtable}

As is shown in Table~\ref{tab: Fault diagnosis experiment}, the RWKDCAE achieves the highest classification accuracy of 100\% on the datasets reconstructed by the AE-DCST. It indicates that the AE-DCST retains all the important features for classification during the compression process. Although the RWKDCAE has a classification accuracy of 100\% on the datasets reconstructed by the AE-DCT, the AE-DCST has a better compression quality score than the AE-DCT as shown in Tables~\ref{tab: Compression experiment on SEU dataset} and~\ref{tab: Compression experiment on Uoc datasets}.
As shown in Fig. \ref{feature}, RWKDCAE models with supervised learning can effectively extract fault features on both testing sets. This also shows that the data compression model can completely retain the important features of the original information.

To validate the function of each layer, we plot the output curves for each layer. In this experiment, we use the SEU testing datasets. As is shown in 
\begin{figure*}[!ht]
	\centering
		\includegraphics[scale=.55]{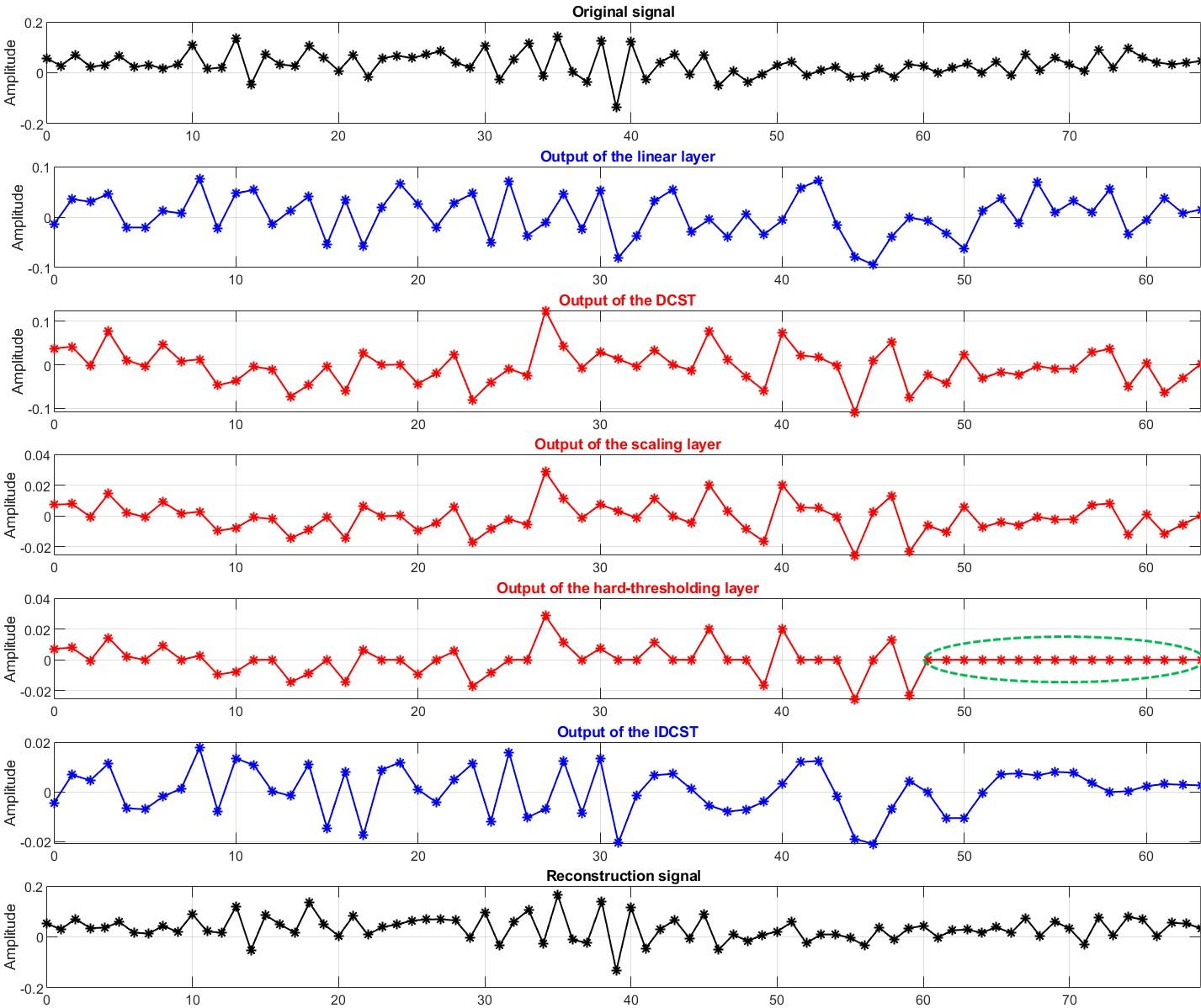}
	\caption{Output of each layer in the AE-DCST.}
	\label{Fig: hard}
\end{figure*}
Fig~\ref{Fig: hard}, the hard-thresholding layer sets small DCST entries to zeros. But, the classification accuracy of the RWKDCAE method on the datasets reconstructed by the proposed method is 100\%. It indicates that the removed information does not contain the key features for classification. Therefore, it is experimentally shown that the hard-thresholding layer removes the redundant data, which is mostly noise. Additionally, it is noticed that the hard-thresholding layer retains the large entries in the low-frequency and high-frequency bands. 
Besides, although there is a large difference between the output of the first linear and the output of the IDCST, the original signal is similar to the reconstructed signal. It indicates the linear in the decoder module has a strong reconstruction ability. Moreover, we observe that the scaling layer scales the output of DCST. Hence, the scaling layer and hard-thresholding layer improve the compression efficiency.

To further verify the generalization ability of the proposed model, we perform a transfer learning study in the fault diagnosis experiments. First, 
the compression models are trained on a small part of the healthy state dataset. Then we test the compression models using the missing tooth, root fault, surface fault, healthy state, and chipped tooth datasets. After receiving five reconstruction datasets, we used the pre-trained RWKDCAE model to perform fault diagnosis tests. 
Here, the pre-trained RWKDCAE model has already learned five fault features in advance. Our target is to check whether these five features are present in the reconstructed datasets or not.
As shown in Table~\ref{table: Transfer fault}, the AE-DCST achieves the best CR, PRD, PRDN, RMS and QS compared with other models in terms of the average results. The detailed results for five gearbox fault types are presented in Tables 6-13 in the supplementary material.
Therefore, the proposed AE-DCST model has a good compression performance on the signals with unlearned defects or out-of-distribution data. 
Moreover, the pre-trained RWKDCAE model has the highest classification accuracy of 99.51\% on the datasets reconstructed by the AE-DCST. 
It indicates that the AE-DCST retains five fault features during the compression process.
This is because the fixed DCST layer has a strong frequency computing capability. It is able to retain the large entries in the low-frequency and high-frequency bands as shown in Fig.~\ref{SEUF}. Hence, it can effectively extract frequency domain features from out-of-distribution data, leading to an enhancement of the generalization ability of the proposed model.

\begin{table}[!ht]
\centering
\caption{Transfering learning fault diagnosis experiment on the SEU datasets.}\label{table: Transfer fault}
%\begin{threeparttable}
\begin{tabular}{c c c c c c c}
\toprule
\multirow{2}{*}{\textbf{Algorithm}}&\textbf{Metrics}  \\ \cmidrule[0.75pt]{2-7}
        \multirow{2}{*}&\textbf{CR}
        &\textbf{PRD}&\textbf{PRDN}&\textbf{RMS}& \textbf{QS}& \textbf{Accuracy}  \\
\midrule
   Multi-layer AE &8.90 & 24.88 & 25.00 & 2.71 & 0.36 &22.20\%  \\
   Sparse AE &9.47 & 24.84 & 24.95 & 2.70 & 0.39&16.50\%\\   
   Variational AE &9.30 & 24.79 & 24.90 & 2.69 & 0.38&19.05\% \\
   HQHTP &9.27 &22.24  &22.34  &2.41   &0.42&74.36\%\\  
   JSNet &8.50 & 28.01 & 28.13 & 3.00 & 0.31&98.02\%\\ 
   Standard DCST &9.16 & 26.43 &26.55 &2.84&0.35&87.69\%\\ 
  AE-DCT &9.42 &22.00 & 22.10 & 2.39&0.43&82.39\%\\ 
  AE-DCST &\textbf{9.59} &\textbf{21.81}& \textbf{21.91} &\textbf{2.36 }&\textbf{0.44}&\textbf{99.51\%}\\ 
\bottomrule
\end{tabular}
 \begin{tablenotes}
        \footnotesize
        % \item This is note content.
\item Note: The models are trained on the health state dataset. Accuracy refers to the classification accuracy of the RWKDCAE method on the datasets reconstructed by the corresponding compression model.
      \end{tablenotes}
\end{table}

\section{Conclusion}
This paper proposes a new asymmetrical autoencoder-type neural network for gearbox sensor data compression. First and foremost, to improve reconstruction accuracy and reduce trainable parameters, a novel sparsifying DCST layer is employed in the traditional autoencoder. The DCST layer with learnable parameters introduces interpretability to the autoencoder because DCT and DCST decorrelate the input data and have the ability to remove the redundant information by quantization. Our proposed layer has learnable parameters and this leads to superior results compared to ordinary DCT and DCST-based schemes with hand-crafted features. Next, the proposed model obtains good results with a small number of training samples. Last but not least, an online data transmission model is developed to improve the compression ratio. The proposed compression model is evaluated on the SEU and the UoC datasets.  
The results show that when compared with other autoencoders and DCT-based methods, the autoencoder with a sparsifying DCST layer is superior in terms of compression efficiency without introducing any diagnostic loss.
The proposed method also achieves a trade-off between compression efficiency and reconstruction accuracy. Thus, it can be applied to gearbox data compression for remote fault diagnosis. The decoder module can be improved with more layers to enhance the ability to reconstruct the compressed signals in the host computer. In future work, the encoder side will be implemented in the chip.

\section*{Acknowledgments}
Xin Zhu was supported by National Science Foundation (NSF) under grant 1934915 and NSF IDEAL 2217023.

%Quantization can be defined as:
%\begin{equation}
%    Q_{i}= \text{sign}(X_{i}\cdot v_{i})(|X_{i}\cdot v_{i}|-T_i)_{+},
%\end{equation}

 \bibliographystyle{elsarticle-num} 
 \bibliography{cas-refs}

\begin{thebibliography}{10}
\expandafter\ifx\csname url\endcsname\relax
  \def\url#1{\texttt{#1}}\fi
\expandafter\ifx\csname urlprefix\endcsname\relax\def\urlprefix{URL }\fi
\expandafter\ifx\csname href\endcsname\relax
  \def\href#1#2{#2} \def\path#1{#1}\fi

\bibitem{compare2022general}
M.~Compare, F.~Antonello, L.~Pinciroli, E.~Zio, A general model for life-cycle cost analysis of condition-based maintenance enabled by phm capabilities, Reliability Engineering \& System Safety 224 (2022) 108499.

\bibitem{jiao2022cycle}
J.~Jiao, J.~Lin, M.~Zhao, K.~Liang, C.~Ding, Cycle-consistent adversarial adaptation network and its application to machine fault diagnosis, Neural Networks 145 (2022) 331--341.

\bibitem{gao2015survey}
Z.~Gao, C.~Cecati, S.~X. Ding, A survey of fault diagnosis and fault-tolerant techniques—part i: Fault diagnosis with model-based and signal-based approaches, IEEE Transactions on Industrial Electronics 62~(6) (2015) 3757--3767.

\bibitem{simani2003model}
S.~Simani, C.~Fantuzzi, R.~J. Patton, Model-based fault diagnosis techniques, in: Model-based Fault Diagnosis in Dynamic Systems Using Identification Techniques, Springer, 2003, pp. 19--60.

\bibitem{zhang2020new}
D.~Zhang, Y.~Chen, F.~Guo, H.~R. Karimi, H.~Dong, Q.~Xuan, A new interpretable learning method for fault diagnosis of rolling bearings, IEEE Transactions on Instrumentation and Measurement 70 (2020) 1--10.

\bibitem{yang2021residual}
D.~Yang, H.~R. Karimi, K.~Sun, Residual wide-kernel deep convolutional auto-encoder for intelligent rotating machinery fault diagnosis with limited samples, Neural Networks 141 (2021) 133--144.

\bibitem{prosvirin2022intelligent}
A.~E. Prosvirin, A.~S. Maliuk, J.-M. Kim, Intelligent rubbing fault identification using multivariate signals and a multivariate one-dimensional convolutional neural network, Expert Systems with Applications 198 (2022) 116868.

\bibitem{lv2022vibration}
Y.~Lv, W.~Zhao, Z.~Zhao, W.~Li, K.~K. Ng, Vibration signal-based early fault prognosis: Status quo and applications, Advanced Engineering Informatics 52 (2022) 101609.

\bibitem{huang2015divide}
Q.~Huang, B.~Tang, L.~Deng, J.~Wang, A divide-and-compress lossless compression scheme for bearing vibration signals in wireless sensor networks, Measurement 67 (2015) 51--60.

\bibitem{guo2013novel}
W.~Guo, W.~T. Peter, A novel signal compression method based on optimal ensemble empirical mode decomposition for bearing vibration signals, Journal of Sound and Vibration 332~(2) (2013) 423--441.

\bibitem{de2015data}
J.~C.~S. de~Souza, T.~M.~L. Assis, B.~C. Pal, Data compression in smart distribution systems via singular value decomposition, IEEE Transactions on Smart Grid 8~(1) (2015) 275--284.

\bibitem{sunil2021bio}
K.~Sunil~Kumar, D.~Shivashankar, K.~Keshavamurthy, Bio-signals compression using auto encoder, Journal of Electrical and Computer Engineering Q 2 (2021) 424--433.

\bibitem{lu2020multi}
H.~Lu, S.~Liu, H.~Wei, J.~Tu, Multi-kernel fuzzy clustering based on auto-encoder for fmri functional network, Expert Systems with Applications 159 (2020) 113513.

\bibitem{wang2020deep}
Y.~Wang, H.~Yang, X.~Yuan, Y.~A. Shardt, C.~Yang, W.~Gui, Deep learning for fault-relevant feature extraction and fault classification with stacked supervised auto-encoder, Journal of Process Control 92 (2020) 79--89.

\bibitem{zemouri2023hydrogenerator}
R.~Zemouri, R.~Ibrahim, A.~Tahan, Hydrogenerator early fault detection: Sparse dictionary learning jointly with the variational autoencoder, Engineering Applications of Artificial Intelligence 120 (2023) 105859.

\bibitem{ko2022new}
J.~U. Ko, K.~Na, J.-S. Oh, J.~Kim, B.~D. Youn, A new auto-encoder-based dynamic threshold to reduce false alarm rate for anomaly detection of steam turbines, Expert Systems with Applications 189 (2022) 116094.

\bibitem{yildirim2018efficient}
O.~Yildirim, R.~San~Tan, U.~R. Acharya, An efficient compression of ecg signals using deep convolutional autoencoders, Cognitive Systems Research 52 (2018) 198--211.

\bibitem{yue2015beyond}
J.~Yue-Hei~Ng, M.~Hausknecht, S.~Vijayanarasimhan, O.~Vinyals, R.~Monga, G.~Toderici, Beyond short snippets: Deep networks for video classification, in: Proceedings of the IEEE Conference on Computer Vision and Pattern Recognition, 2015, pp. 4694--4702.

\bibitem{wallace1991jpeg}
G.~K. Wallace, The jpeg still picture compression standard, Communications of the ACM 34~(4) (1991) 30--44.

\bibitem{taubman2002jpeg2000}
D.~S. Taubman, M.~W. Marcellin, M.~Rabbani, Jpeg2000: Image compression fundamentals, standards and practice, Journal of Electronic Imaging 11~(2) (2002) 286--287.

\bibitem{chen2020jsnet}
B.~Chen, Y.~Wu, G.~Coatrieux, X.~Chen, Y.~Zheng, Jsnet: a simulation network of jpeg lossy compression and restoration for robust image watermarking against jpeg attack, Computer Vision and Image Understanding 197 (2020) 103015.

\bibitem{aydin1991ecg}
M.~Aydin, A.~E. {\c{C}}etin, H.~K{\"o}ymen, Ecg data compression by sub-band coding, Electronics Letters 27~(4) (1991) 359--360.

\bibitem{jha2018electrocardiogram}
C.~K. Jha, M.~H. Kolekar, Electrocardiogram data compression using dct based discrete orthogonal stockwell transform, Biomedical Signal Processing and Control 46 (2018) 174--181.

\bibitem{243411}
A.~E. Cetin, H.~Koymen, M.~Aydin, Multichannel ecg data compression by multirate signal processing and transform domain coding techniques, IEEE Transactions on Biomedical Engineering 40~(5) (1993) 495--499.
\newblock \href {https://doi.org/10.1109/10.243411} {\path{doi:10.1109/10.243411}}.

\bibitem{cetin2006compression}
A.~E. Cetin, H.~K{\"o}ymen, Compression of digital biomedical signals, in: The Biomedical Engineering Handbook: Medical Devices and Systems, CRC Press, 2006, pp. 3--1.

\bibitem{cetin1994coding}
A.~E. Cetin, A.~Tewfik, Y.~Yardimci, Coding of ecg signals by wavelet transform extrema, in: Proceedings of IEEE-SP International Symposium on Time-Frequency and Time-Scale Analysis, IEEE, 1994, pp. 544--547.

\bibitem{978}
J.~Ladan, E.~R. Vrscay, The discrete orthonormal stockwell transform and variations, with applications to image compression, in: Image Analysis and Recognition, Springer Berlin Heidelberg, Berlin, Heidelberg, 2013, pp. 235--244.

\bibitem{shao2018highly}
S.~Shao, S.~McAleer, R.~Yan, P.~Baldi, Highly accurate machine fault diagnosis using deep transfer learning, IEEE Transactions on Industrial Informatics 15~(4) (2018) 2446--2455.

\bibitem{cao2018preprocessing}
P.~Cao, S.~Zhang, J.~Tang, Preprocessing-free gear fault diagnosis using small datasets with deep convolutional neural network-based transfer learning, IEEE Access 6 (2018) 26241--26253.

\bibitem{tan2008performance}
C.~C. Tan, C.~Eswaran, Performance comparison of three types of autoencoder neural networks, in: 2008 Second Asia International Conference on Modelling \& Simulation (AMS), IEEE, 2008, pp. 213--218.

\bibitem{ahmed1974discrete}
N.~Ahmed, T.~Natarajan, K.~R. Rao, Discrete cosine transform, IEEE Transactions on Computers 100~(1) (1974) 90--93.

\bibitem{qtable}
A.~Efros, Lossy image compression (jpeg), \url{https://cs.brown.edu/courses/csci1950-g/lectures/9/DCT.ppt} (2010).

\bibitem{wang2019novel}
F.~Wang, Q.~Ma, W.~Liu, S.~Chang, H.~Wang, J.~He, Q.~Huang, A novel ecg signal compression method using spindle convolutional auto-encoder, Computer Methods and Programs in Biomedicine 175 (2019) 139--150.

\bibitem{chen2021feature}
H.~Chen, X.~He, H.~Yang, L.~Qing, Q.~Teng, A feature-enriched deep convolutional neural network for jpeg image compression artifacts reduction and its applications, IEEE Transactions on Neural Networks and Learning Systems 33~(1) (2021) 430--444.

\bibitem{618009}
P.~Noll, Mpeg digital audio coding, IEEE Signal Processing Magazine 14~(5) (1997) 59--81.
\newblock \href {https://doi.org/10.1109/79.618009} {\path{doi:10.1109/79.618009}}.

\bibitem{pan2022dct}
H.~Pan, X.~Zhu, S.~Atici, A.~E. Cetin, Dct perceptron layer: A transform domain approach for convolution layer, arXiv preprint arXiv:2211.08577 (2022).

\bibitem{park2003m}
H.~Park, Y.~Park, S.-K. Oh, L/m-fold image resizing in block-dct domain using symmetric convolution, IEEE Transactions on Image Processing 12~(9) (2003) 1016--1034.

\bibitem{jiang2019dct}
W.~Jiang, Z.~Wang, J.~S. Jin, Y.~Han, M.~Sun, Dct--cnn-based classification method for the gongbi and xieyi techniques of chinese ink-wash paintings, Neurocomputing 330 (2019) 280--286.

\bibitem{mallat1999wavelet}
S.~Mallat, A wavelet tour of signal processing, Elsevier, 1999.

\bibitem{donoho1995noising}
D.~L. Donoho, De-noising by soft-thresholding, IEEE Transactions on Information Theory 41~(3) (1995) 613--627.

\bibitem{ng2011sparse}
A.~Ng, et~al., Sparse autoencoder, CS294A Lecture Notes 72~(2011) (2011) 1--19.

\bibitem{akhter2010ecg}
S.~Akhter, M.~Haque, Ecg comptression using run length encoding, in: 2010 18th European Signal Processing Conference, IEEE, 2010, pp. 1645--1649.

\bibitem{chen1977fast}
W.-H. Chen, C.~Smith, S.~Fralick, A fast computational algorithm for the discrete cosine transform, IEEE Transactions on communications 25~(9) (1977) 1004--1009.

\bibitem{sharma2010compression}
M.~Sharma, et~al., Compression using huffman coding, IJCSNS International Journal of Computer Science and Network Security 10~(5) (2010) 133--141.

\bibitem{seu}
S.~Shao, S.~McAleer, R.~Yan, P.~Baldi, Mechanical-datasets, \url{https://github.com/cathysiyu/Mechanical-datasets} (2018).

\bibitem{uoc}
P.~Cao, S.~Zhang, J.~Tang, Gear fault data, \url{https://doi.org/10.6084/m9.figshare.6127874.v1} (2018).

\bibitem{zhao2020deep}
Z.~Zhao, T.~Li, J.~Wu, C.~Sun, S.~Wang, R.~Yan, X.~Chen, Deep learning algorithms for rotating machinery intelligent diagnosis: An open source benchmark study, ISA Transactions 107 (2020) 224--255.

\bibitem{loshchilov2017decoupled}
I.~Loshchilov, F.~Hutter, Decoupled weight decay regularization, arXiv preprint arXiv:1711.05101 (2017).

\bibitem{pan2023hybrid}
H.~Pan, X.~Zhu, S.~F. Atici, A.~Cetin, A hybrid quantum-classical approach based on the hadamard transform for the convolutional layer, in: International Conference on Machine Learning, PMLR, 2023, pp. 26891--26903.

\bibitem{oliveira2023early}
A.~Oliveira-Filho, R.~Zemouri, P.~Cambron, A.~Tahan, Early detection and diagnosis of wind turbine abnormal conditions using an interpretable supervised variational autoencoder model, Energies 16~(12) (2023) 4544.

\end{thebibliography}





% \end{thebibliography}
\end{document}